\documentclass[preprint,12pt]{elsarticle}

\usepackage{amssymb}
\usepackage{amsmath}

\usepackage{algorithm}
\usepackage{algpseudocode}

\journal{I. J. for Numerical and Analytical Methods in Geomechanics}

\begin{document}

\begin{frontmatter}
    
    
    
    \title{A POD-TANN approach for the multiscale modeling of materials and macroelement derivation in geomechanics}
    
    
    \author[inst1]{Giovanni Piunno}
    
    \affiliation[inst1]{organization={Department of Civil and Environmental Engineering, Politecnico di Milano},
                addressline={Piazza leonardo da Vinci, 32}, 
                city={Milan},
                postcode={20133}, 
                state={Italy},
                country={Italy}}
    
    \author[inst2]{Ioannis Stefanou}
    \author[inst1]{Cristina Jommi}
    
    \affiliation[inst2]{organization={Nantes Université, Ecole Centrale Nantes, CNRS,
    Institut de Recherche en Génie Civil et Mécanique (GeM)},
                addressline={1 rue de la Noë BP 92101}, 
                city={Nantes},
                postcode={44321}, 
                state={Pays de la Loire},
                country={France}}
    
    \begin{abstract}
    This paper introduces a novel approach that combines Proper Orthogonal Decomposition (POD) with Thermodynamics-based Artificial Neural Networks (TANN) to capture the macroscopic behavior of complex inelastic systems and derive macroelements in geomechanics. 

The methodology leverages POD to extract macroscopic Internal State Variables from microscopic state information, thereby enriching the macroscopic state description used to train an energy potential network within the TANN framework. The thermodynamic consistency provided by TANN, combined with the hierarchical nature of POD, allows to reproduce complex, non-linear inelastic material behaviors as well as macroscopic geomechanical systems responses.

The approach is validated through applications of increasing complexity, demonstrating its capability to reproduce high-fidelity simulation data. The applications proposed include the homogenization of continuous inelastic representative unit cells and the derivation of a macroelement for a geotechnical system involving a monopile in a clay layer subjected to horizontal loading. Eventually, the projection operators directly obtained via POD, are exploit to easily reconstruct the microscopic fields.

The results indicate that the POD-TANN approach not only offers accuracy in reproducing the studied constitutive responses, but also reduces computational costs, making it a practical tool for the multiscale modeling of heterogeneous inelastic geomechanical systems.
    \end{abstract}
    
    
    

\end{frontmatter}


\section{Introduction}\label{S1}
The study of matter through multi-scale analysis has revealed that what appears homogeneous at the macroscopic level is actually composed of numerous interacting components at the microscopic level, giving rise to various macroscopic effects. Historically, engineering approaches have simplified the description of the macroscopic behavior of complex systems, often relying on phenomenological models that replicate experimentally observed behaviors. While successful, these models fail to account for microscopic processes and interactions, limiting their accuracy and predictive capability.

The complexity and variability of microscopic interactions, combined with the increasing availability of high-dimensional data, have motivated the exploration of machine learning (ML) techniques, which offer powerful tools for capturing intricate relationships that traditional methods may overlook.

There are two main challenges in establishing a ML-based homogenization procedure. Firstly, the obtained model at the macroscale must comply with basic physical laws, a general requirement not specific to ML techniques adopted to derive it. This necessitates incorporating such laws into the structure of the ML tools. Recently, neural network-based approaches have been proposed that incorporate basic physical laws into the model structure by design (see \cite{raissi2019physics}, \cite{ibanez2018manifold}, \cite{cueto2022thermodynamics}, \cite{flaschel2023automated}, among others). Secondly, the multi-scale problem must account for the properties and responses of the constituents at the microscopic level. Examples of ML-based strategies for accelerating multi-scale analyses can be found in \cite{rocha2023deepbnd}, \cite{yin2022interfacing}.

Resolving the coupled problem described above is an open and active research topic. In this study, a Thermodynamics-based Artificial Neural Networks (TANN) formulation \cite{Masi_2021_a}, \cite{Masi_2022}, \cite{Masi_ETANN} is adopted, and a new and more systematic method for identifying the Internal State Variables (ISVs) of micro-structured inelastic systems is proposed, based on Proper Orthogonal Decomposition (POD). Specifically, POD is integrated into the TANN workflow to learn the energy potential of the macroscopic homogenized system.

Traditionally, POD has been used to extract the dominant spatial patterns (modes) of a high-dimensional system by performing a singular value decomposition (SVD) on the snapshot matrix of the system's state variables. The resulting modes are then used to construct a low-dimensional Reduced Order Model for the system by projecting the PDEs describing the system's evolution onto the modes and solving the reduced equations using Galerkin methods \cite{lumley_1967}, \cite{holmes2012turbulence}, \cite{brunton2022data}, \cite{sampaio2007remarks}, \cite{kerschen2002physical}.

In this work, POD is applied to a set of microscopic state variables to extract the internal state variables at the macroscopic scale as a result of dimensionality reduction. Unlike other techniques, POD offers an unsupervised identification methodology that is hierarchical, preserving maximal information in the reduced space given the dimension of the embedding. This feature is exploited in the paper to select a sufficiently large space of ISVs for training the energy network, a neural network used to learn the macroscopic (Helmholtz) free energy of the composite. The hierarchical nature of POD, along with its robustness and efficiency, significantly reduces the computational cost by decoupling the ISV identification from the training of the energy network.

The proposed method bears some similarities to the analytical Nonuniform Transformation Field Analysis (NTFA) method by Michel and Suquet \cite{Michel2003}, \cite{Michel2004}, \cite{Michel2010}, \cite{Michel2016}, based on the pioneering work of Dvorak \cite{Dvorak_1992}. The NTFA theory focuses on composites made of Generalized Standard Materials (GSM), as defined by \cite{halphen1975materiaux}, and aims to determine the reduced potentials of the composite, i.e., those describing the macroscopic effective behavior, from the knowledge of the constituent models \cite{suquet1985local}. At the core of the NTFA approach, similar to what is proposed in this paper, is the selection of a reduced basis to perform a model reduction of the local internal variables of the nonlinear constituents by projection. However, unlike the NTFA method, the methodology proposed here can address a more general class of materials and provides a unified workflow to derive the homogenized model, relying on the combined use of POD and thermodynamics-based artificial neural networks.

POD is also used as a practical and efficient tool for bridging the micro- and macro scales bidirectionally. In the upscaling direction, it characterizes the macroscopic state from microscopic information. In the downscaling direction, POD reconstructs the microscopic fields with a given accuracy defined by the reconstruction error. The modes used for projection (upward) and reconstruction (downward) are computed in a pre-processing step once and for all, prior to the training of the TANN energy network. This approach is extremely time-saving compared to alternative methods like autoencoders used in \cite{Masi_2022}, which require coupled training with the TANN network each time.

The examples of applications presented in this study are of two types. First, the multi-scale homogenization problem of elasto-plastic heterogeneous materials with continuous microstructure is addressed. Second, the same POD-TANN framework is used to derive a macroelement featuring a single degree of freedom from a complex three-dimensional FE model of a geotechnical system. This latter example is focused on a monopile in a clay layer subjected to horizontal loading.

The paper is divided into four sections. Section \ref{S2} reviews the theoretical framework. In Section \ref{S3}, TANNs are discussed, and the proposed POD-TANN approach is outlined, starting with a brief review of the POD method to explain its use for the multi-scale problem. Section \ref{S4} presents applications of the methodology. Section \ref{S5} offers the conclusions drawn from the research.

\section{Theoretical background}\label{S2}

Constitutive equations maps system-specific responses to a given set of external actions under a given set of constraints. In continuum mechanics, constitutive equations also have a precise role in the mathematical formulations of the system of differential field equations of the studied body. Due to the unbalanced number of unknown fields, which is larger than the number of field equations derived from balance and conservation laws, the latter describing the physics of the problem, constitutive equations are needed to achieve the closure of the system. 

Here the thermodynamic framework introduced by Coleman and Gurtin \cite{Coleman1967ThermodynamicsWI} is considered, which provides the necessary mathematical formalism for the constitutive description of a large class of materials (see \cite{Muschik_2008} for an extensive review). 

Coleman and Gurtin have based their formulation on the definition of a state spaces, $\mathcal{S}(t)$, including basic quantities, such as the strains or stresses, enriched with an additional set of Internal State Variable ($ISVs$), defined at the current time $t$, denoted by $\boldsymbol{z}$. The ISVs evolve according to a set of additional equations, called evolution laws.

At the core of the Coleman and Gurtin's approach is the use of the second law, in the form of the Clausius-Duhem inequality. This is exploited, together with the first law, to obtain restrictions on the form of the constitutive equations. A general review of the theory may be found in \cite{Masi_2021_a}. For the sake of simplicity, isothermal processes and infinitesimal strain regime will be considered in this work. The relevant quantities are introduced as follows: $\boldsymbol{\epsilon}=Sym\left(\nabla \boldsymbol{u}\right)$, $p^{in}=\boldsymbol{\sigma}:\dot{\boldsymbol{\epsilon}}$, with $\boldsymbol{u}$ being the displacement field, $\boldsymbol{\sigma}$ the Cauchy stress tensor, $\boldsymbol{\epsilon}$ the infinitesimal strain tensor, $p^{in}$ the internal power, $\nabla$ the gradient operator and $Sym(\bullet)$ the symmetric part operator. 

In this simplified setting, the thermodynamic restrictions derived from the exploited Clausius-Duhem inequality are provided as follows:
\begin{equation}\label{EQ4}
    \begin{split}
        &(I \quad \text{Law}) \quad p^{in} = \dot{\psi} + d;\\
        &(II\quad \text{Law}) \quad d\geq0;\\
        &(\text{Constraints}) \quad \boldsymbol{\sigma} - \frac{\partial\hat{\psi}}{\partial\boldsymbol{\epsilon}}=0, \quad d = - \frac{\partial \hat{\psi}}{\partial \boldsymbol{z}} \cdot \dot{\boldsymbol{z}} \geq 0.
    \end{split}
\end{equation}
In the above expressions, $\psi=\hat{\psi}(\boldsymbol{\epsilon}, \boldsymbol{z})$ is the Helmholtz free energy density, $d$ is the rate of mechanical dissipation. 
Furthermore, depending on the setting chosen by the modeler, Fenchel-Legendre transformations of the potential energy density are introduced so to switch from dependent and independent static and kinematic variables. To aid the formulation used in the following secitons, the transform between the Helmholtz and Gibbs free energy densities is recalled:
\begin{equation}\label{EQ5}
    \begin{split}
        &\psi - \phi = \boldsymbol{\sigma}:\boldsymbol{\epsilon};\\
        &\boldsymbol{\epsilon} + \frac{\partial\hat{\phi}}{\partial\boldsymbol{\sigma}}=0, \quad d = - \frac{\partial \hat{\phi}}{\partial \boldsymbol{z}} \cdot \dot{\boldsymbol{z}} \geq 0;
    \end{split}
\end{equation}
where $\phi=\hat{\phi}(\boldsymbol{\sigma}, \boldsymbol{z})$ has been introduced for the Gibbs free energy density. 

\subsection{Formulation for multiscale homogenization problems}\label{S2.1}
The formal derivation of the constitutive restrictions obtained in \cite{Coleman1967ThermodynamicsWI} is retained when a macroscopic constitutive equation is derived by homogenization of a microscopic system under prescribed boundary conditions.
This has been demonstrated in \cite{Masi_2022}, considering the framework of the asymptotic (espansion) homogenization (AEH) theory, in incremental formulation, see, e.g., \cite{miehe2002strain}, \cite{Pinho_2009}, \cite{bakhvalov2012homogenisation}. 

The theory considers an heterogeneous body $\mathcal{B}$, with volume $\Omega$, characterized by a microstructure made of a spatially periodic distribution of a Representative Unit Cell (RUC), with volume $\mathcal{Y}$. The average value of a quantity in $\mathcal{Y}$ is denoted as:
\begin{equation}\label{EQ6}
    \left\langle\bullet\right\rangle=\frac{1}{\left| Y \right|} \int\sb{Y}{(\bullet)d\boldsymbol{y}},
\end{equation}
where any point in the RUC is localized using the local coordinate vector $\boldsymbol{y}$, while the RUC is macroscopically localized in the body $\mathcal{B}$ by the macroscopic coordinate vector $\boldsymbol{x}$.\\
All the fields in $\Omega$ show $\mathcal{Y}$-periodicity, see \cite{bakhvalov2012homogenisation}. This property is indicated with the superscript $\bullet^e$. From the solution of the auxiliary problem on the RUC (see \cite{miehe2002strain}), subjected to periodic displacements and anti-periodic traction vectors, the volume average macroscopic stress increments resulting from the imposed macroscopic strain increments can be found. Constitutive equations linking independent and dependent state variables at the macroscale, as the macroscopic stress and strain, still have to satisfy thermodynamic restrictions introduced in equations \ref{EQ4}. Therefore, the volume averaged expressions of the first and second laws at the macroscopic level can be obtained, and are valid for any material point of the homogenized body $\mathcal{B}_H$. The expressions are formally identical to those reported in equation \ref{EQ4} and, exploiting the Colemn and Gurtin's procedure, lead to following macroscopic restrictions:
\begin{equation}\label{EQ7}
    \boldsymbol{\Sigma} - \frac{\partial\hat{\Psi}}{\partial\boldsymbol{E}}=0, \quad D = - \frac{\partial \hat{\Psi}}{\partial \boldsymbol{Z}} \cdot \dot{\boldsymbol{Z}} \geq 0.
\end{equation}
in which the following volume average quantities are defined: the macroscopic Cauchy stress tensor, $\boldsymbol{\Sigma}(\boldsymbol{x})\triangleq\left\langle\boldsymbol{\sigma}^{e}(\boldsymbol{x}, \boldsymbol{y})\right\rangle$, the macroscopic small strain tensor $\boldsymbol{E}(\boldsymbol{x}, \boldsymbol{y})\triangleq\left\langle\boldsymbol{\epsilon}^{e}(\boldsymbol{x}, \boldsymbol{y})\right\rangle$, the macroscopic Helmholtz free energy density 
$\Psi(\boldsymbol{x}, \boldsymbol{y})\triangleq\left\langle\psi^{e}(\boldsymbol{x}, \boldsymbol{y})\right\rangle$, the macroscopic rate of mechanical dissipation
$D(\boldsymbol{x}, \boldsymbol{y})\triangleq\left\langle d^{e}(\boldsymbol{x}, \boldsymbol{y})\right\rangle$, the macroscopic vector of internal state variables $\boldsymbol{\mathcal{Z}}$. 

\subsection{Formulation for macroelement definition}\label{S2.2}
In the previous sections, strategies were discussed for modeling the mechanical behavior of materials and obtaining a homogenized representation of the behavior of heterogeneous microstructured media, such as those typically encountered in geotechnics. However, in geotechnical engineering, it is challenging to distinguish the behavior of a macroscopic system, such as soil or a natural deposit, from that of the anthropogenic structures interacting with it. In other words, it is often necessary to study what is commonly referred to as soil-structure interaction (SSI).

SSI problems remain an unresolved issue in many geotechnical applications and pose a significant challenge for designers, as standard design approaches struggle to address such complex and multiscale problems. Moreover, commercial numerical codes, although increasingly accurate, often demand substantial computational resources and advanced theoretical knowledge, making them impractical as standard design tools, particularly in the preliminary stages of the design process.

In this context, where direct and time-efficient applicability to real engineering problems is crucial, macroelement approaches frequently offer a viable alternative modeling option. The key is to describe the SSI in terms of generalized stress and strain variables. This involves lumping stress variables, such as vertical and horizontal forces or moments, and soil strain fields into kinematic variables, representing structural displacements or rotations relative to the far-field boundary.

Macroelements allow for the problem to be upscaled from the local scale where numerical integration over the spatial domain is required, provided a proper constitutive relationship is assigned to each representative elementary volume or unit cell, equilibrium and compatibility equations are satisfied, and boundary and initial conditions are specified to the macro scale of the structure and its surrounding soil. In other words, an ad hoc “generalized” constitutive relationship is introduced, relating the aforementioned generalized stress and strain variables. The structure and the interacting soil are then considered as a single “macro” element, whose behavior is fully described by a limited number of degrees of freedom \cite{galli2020macroelement}.

The important advantage of the macroelement approach is the possibility of exploiting pre-existing mechanical and thermodynamical theories defined at the local material scale to the scale of the whole engineering system, introducing a certain degree of empiricism in the modeler choice of the set of generalized static and kinematic variables. By denoting with $\boldsymbol{F}$ and $\boldsymbol{u}$ the work-conjugate set of generalized forces and displacements, the first principle can be written as follows:
\begin{equation}\label{EQ8}
    \dot{f} = \boldsymbol{F}\cdot\dot{\boldsymbol{u}} - D,
\end{equation}
having introduced the Helmoltz energy of the system, $f$. Using the same arguments mentioned above and exploiting the Legendre transform to formulate the problem using the force-driven Gibbs energy, $g=\hat{g}(\boldsymbol{F}, \boldsymbol{Z})$, the thermodynamic restriction of the constitutive model of the macro system can be written as follows:
\begin{equation}\label{EQ9}
    \boldsymbol{u} - \frac{\partial\hat{g}}{\partial\boldsymbol{F}}=0, \quad D = - \frac{\partial \hat{g}}{\partial \boldsymbol{Z}} \cdot \dot{\boldsymbol{Z}} \geq 0,
\end{equation}
where for the mechanical rate of dissipation and the set of macroscopic internal state variables, the symbols $D$ and $\boldsymbol{Z}$ have been used, in analogy to what done for the rate of dissipation and ISV vector of the homogenized RUC. 

\section{Thermodynamic-based Artificial Neural Networks for the multiscale problem}\label{S3}
Relying on the theoretical background presented in Section \ref{S2}, Masi et al. \cite{Masi_2021_a} have developed the so called Thermodynamics-based Artificial Neural Networks - TANN, to learn the constitutive behavior of inelastic homogeneous materials. The initial TANN formulation has been further extended from homogeneity, to discrete heterogeneous micro-structured inelastic materials in \cite{Masi_2022}. 

At the core of the Coleman and Gurtin's theory is the learning of two functions: the Helmholtz free energy density function and the evolution law of the internal state variables. Therefore, the TANN consists of two networks: one is used to learn the potential energy function from data; the second one is used to learn the evolution law from data. The balance of energy in \ref{EQ4} is imposed during training by construction, differentiating the energy network's output with respect to its strain input, so to get the stress tensor exploiting auto-differentiation, see \cite{baydin2018automatic}. The non-negativeness of the rate of mechanical dissipation is learned from data. This is achieved including in the loss function of the energy network a regularization term in the rate of dissipation.

More specifically, the energy network inputs are the state of the material $\mathcal{S}$ at time $t$, namely $\boldsymbol{\epsilon}(t)$, $\boldsymbol{z}(t)$. The TANN is trained to output the Helmoltz free-energy at time $t$, $\psi = f^{\psi}(\mathcal{S})$. The stress at time $t$ is obtained as the partial derivative of $f^{\psi}$ with respect to the infinitesimal strains, $\boldsymbol{\sigma}=\partial_{\boldsymbol{\epsilon}} f^{\psi}$, and the mechanical dissipation rate as follows, $d = -\partial_{\boldsymbol{z}} f^{\psi}\cdot \dot{\boldsymbol{z}}$. The regularization term is included in the training loss to penalize negative predictions of the rate of dissipation. It is defined as $\{d\} = Relu\left( - \left(-\partial_{\boldsymbol{z}} f^{\psi}\cdot \dot{\boldsymbol{z}} \right) \right)$, with $Relu\left(\bullet \right) = \{\bullet, \quad if \quad \bullet >0;\quad 0, \quad otherwise\}$. 

The loss function, $\mathcal{L}$, is therefore the sum of four weighted terms:
\begin{equation}\label{EQ10}
    \mathcal{L} = w^{\psi}\ell^{\psi} + w^{\boldsymbol{\sigma}}\ell^{\boldsymbol{\sigma}} + w^{d}\ell^{d} + w^{\{d\}}\ell^{\{d\}}.
\end{equation}
The weights, $w^i$ with $i$ in $[\psi, \boldsymbol{\sigma}, d, \{d\}]$, are used to assess the relative output influence on the training process; the training losses, $\ell^i$, are computed using a metric of the error between the output and the training data. 

\subsection{Discovery of internal state variables using model reduction}\label{S3.1}
Following the adopted thermodynamic framework, the energy network can be trained once the state space of the material is completely defined. For macroscopic micro-structured systems it is not straightforward to fully characterize the state space, and defining the vector of ISV. In \cite{Masi_2022}, the problem is addressed by means of dimensionality reduction techniques on the so called Internal Coordinates (IC), $\boldsymbol{\xi}$, i.e. the set of all the quantities describing the material state at the microscopic scale.

It is remarked that \textit{Encoding} refers to the computation of a reduced and parametrized representation of a state, whereas \textit{Decoding} or reconstruction describes the computation of the actual state from the parametrized representation.

In \cite{Masi_2022}, autoencoders are used to learn an unsupervised lower-dimensional representation of the IC, serving as the ISV vector at the macroscale. In this work, the Proper Orthogonal Decomposition (POD) is used instead as a simplified, jet effective tool for the unsupervised identification of macroscopic ISVs from a set of microscopic Internal Coordinates. The POD method can be seen as an encoder/decoder based on linear projections, thus can be recoverd as a special case of an autoencoder \cite{fresca2022pod}. However, several reasons motivate the use of the classical POD method in this work.
\begin{itemize}
    \item By definition, the set of internal state variables must accurately represent the processes occurring at the microscale. However, the concept of representativeness requires careful clarification. A set of internal variables is considered representative if, in conjunction with the macroscopic strain state, it can provide a state space sufficiently large to establish a one-to-one relationship with the Helmholtz free energy density through a state function. In the context of thermodynamic-based artificial neural networks, this entails identifying a set that minimizes the loss function of the energy network to a specified threshold, while in POD is directly accessible by means of the Singular Value Decomposition of the snapshot matrix.
    \item The Proper Orthogonal Decomposition is a hierarchical dimensionality reduction technique designed to preserve as much information as possible within a reduced dimensional space. Achieving a predefined threshold of training loss involves increasing the number of considered POD components, ensuring the accuracy of the micro-structural processes is maintained. 
    \item The maximum representativeness achieved through the hierarchical sorting of POD modes provides a significant advantage: it eliminates the need to couple POD regression with the training of the energy network. In contrast, when using an encoder, the reduced dimensionality of the output does not necessarily preserve the maximum amount of information from the original set. The training of an autoencoder is only ensuring that the composition of the nonlinear mappings of the encoder and decoder results in the identity operator. This necessitates coupling the training of the encoder within the TANN workflow to ensure the thermodynamic consistency of the dimensionality reduction, as detailed in \cite{Masi_2022}. Consequently, POD, being less computationally expensive, significantly reduces the computational cost associated with training the energy network.
    \item The POD is a linear method that involves matrix multiplications, ensuring the existence of an inverse function. Consequently, there is no need to train an additional decoder to reconstruct the microscopic fields collected in the internal coordinates from the internal state variables vector.
    \item The POD algorithm is fast, robust, and efficient, and does not require the finetuning of complicated hyperparameters, as in the case of autoencoders.
\end{itemize}

In the proposed POD-TANN framework, the POD is employed to reduce the dimensionality of the internal coordinates, $\boldsymbol{\xi}$, which are responsible for tracking the irreversible processes occurring at the microscale. By applying POD to the snapshot matrix $\boldsymbol{\Xi}$, constructed from successive snapshots of $\boldsymbol{\xi}$, the ICs are approximated using a set of optimal spatial modes, $\boldsymbol{\Phi}$, and time-varying coefficients, $\boldsymbol{\mathcal{Z}}$.

Mathematically, the field of internal coordinates, $\boldsymbol{\xi}$, can be expressed as:

\begin{equation}
    \boldsymbol{\xi}(\boldsymbol{x}, t)=\sum_{k=1}^{r}z_{k}(t)\boldsymbol{\Phi}_k(\boldsymbol{x}) = \boldsymbol{\Phi}(\boldsymbol{x})\boldsymbol{Z}(t),
\end{equation}

where $r$ is the number of retained POD modes, $\boldsymbol{\Phi}_k(\boldsymbol{x})$ are the time-independent spatial modes, and $\boldsymbol{Z}(t)$ is the collection of vectors of time-dependent POD coefficients, representing the reduced-order internal state variables (ISVs) at each time step.

The optimal POD basis, $\boldsymbol{\Phi}$, is derived through the Singular Value Decomposition (SVD) of the snapshot matrix $\boldsymbol{\Xi}$, which organizes the snapshots of the internal coordinate field. Through SVD, $\boldsymbol{\Xi}$ is approximated and decomposed as:
\begin{equation}
    \boldsymbol{\Xi}\approx\tilde{\boldsymbol{\Xi}}=\tilde{\boldsymbol{U}}\tilde{\boldsymbol{S}}\tilde{\boldsymbol{V}}^{*},
\end{equation}
where $\tilde{\boldsymbol{U}}$ contains the first $r$ time-invariant spatial modes (POD modes) of $\boldsymbol{U}$, $\tilde{\boldsymbol{S}}$ holds the retained singular values, and $\tilde{\boldsymbol{V}}^{*}$ contains the time coefficients, with $*$ indicating the transposed-conjugate. By retaining only the dominant $r$ modes, the dimensionality of the ICs is reduced, yielding a compact yet representative set of ISVs, $\boldsymbol{\mathcal{Z}}$, that describe the system's evolution. $\boldsymbol{\mathcal{Z}}$ is obtained projecting $\boldsymbol{\xi}$ onto the basis defined by the orthogonal retained modes, namely $\boldsymbol{\mathcal{Z}}=\tilde{\boldsymbol{U}}^{*}\boldsymbol{\xi}$ (see Appendix A for a detailed derivation).
Exploiting the time-invariance of the POD-modes, the rate of change of the ISV are also easily found from the rate of change of the ICs, as follows:
\begin{equation}
    \dot{\boldsymbol{\mathcal{Z}}}=\frac{d}{dt}(\tilde{\boldsymbol{U}}^{*}\boldsymbol{\xi})=\tilde{\boldsymbol{U}}^{*}\dot{\boldsymbol{\xi}}.
\end{equation}

\subsubsection{State definition for the homogenization problem}\label{S3.1.1}
\begin{figure}
\centering
    \includegraphics[width=0.8\textwidth]{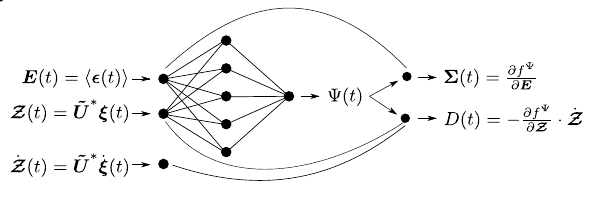}
\caption{Sketch of the energy network of the POD-aided TANN used for the discovery of the macroscopic Helmholtz free energy density function of the homogenized micro-structured medium.}
\label{FIG1}
\end{figure}

The integration of the POD within the TANN framework is achieved performing the SVD onto a snapshot matrix of the IC, resulting from the collection of the strain paths used to create a database. Once the eigenbasis is found, the ISV vector is simply found by projection of the IC onto the POD modes. Considering the macroscopic state of the homogenized continuum, described in section \ref{S2.1}, 
the following can be written:
\begin{equation}\label{EQ14}
    \mathcal{S}=\left[\boldsymbol{E}, \boldsymbol{\mathcal{Z}}\right] = \left[\boldsymbol{E}, \boldsymbol{\tilde{U}}^* \boldsymbol{\xi}\right].
\end{equation}
Eventually, the TANN workflow can be used without any modification at the macroscale, once the above defined volume averaged quantities are considered. Figure \ref{FIG1} shows a sketch of the energy network's architecture at the macroscale, adopted in section \ref{S4}. 

\subsubsection{State definition for the macroelement}\label{S3.1.2}
Adopting an equivalent approach on a snapshot matrix of the IC of the macroscopic system, the state space for the macroelement can be defined as follows:
\begin{equation}\label{EQ15}
    \mathcal{S}=\left[\boldsymbol{U}, \boldsymbol{\mathcal{Z}}\right] = \left[\boldsymbol{U}, \boldsymbol{\tilde{U}}^* \boldsymbol{\xi}\right],
\end{equation}
or, aleternatively:
\begin{equation}\label{EQ16}
    \mathcal{S}=\left[\boldsymbol{F}, \boldsymbol{\mathcal{Z}}\right] = \left[\boldsymbol{F}, \boldsymbol{\tilde{U}}^* \boldsymbol{\xi}\right],
\end{equation}
depending if a force- or displacement-based approach is adopted.

\subsection{Energy reconstruction error}\label{S3.2}
The ability to decouple energy network training from the identification of the internal state variables (ISV) is attributed to the hierarchical nature of the Proper Orthogonal Decomposition (POD). However, determining the exact number of POD modes required for a low-rank approximation of $\boldsymbol{\xi}$ is not straightforward. This necessitates exploring the possibility of identifying a sufficient number of POD modes that can ensure the training of the energy network meets a predetermined error bound.

In practical applications, the number of retained POD modes is typically chosen based on a threshold for the cumulative sum of the normalized singular values, usually set at 0.95 or higher. However, this choice often lacks a solid theoretical justification \cite{brunton2022data}. In this context, \cite{gavish2014optimal} have proposed a mathematical procedure to determine the optimal hard threshold for SVD truncation, assuming that the high-rank matrix has a low-rank structure contaminated with Gaussian white noise.

This work adopts a different approach. In particular, the number of POD modes is chosen by requiring a predefined error tolerance between the exact, $\Psi$, and the reconstructed, $\bar{\Psi}$, free energy of the system.

Starting from the vector of collected IC, $\boldsymbol{\xi}$, it is possible to apply the SVD and compute a reconstructed IC vector, $\bar{\boldsymbol{\xi}} = \boldsymbol{\bar{U}}\boldsymbol{\bar{U}}^{*}\boldsymbol{\xi}=\bar{\boldsymbol{U}}\bar{\boldsymbol{S}}\bar{\boldsymbol{V}}^{*}$, function of the given number of POD modes considered. From $\bar{\boldsymbol{\xi}}$, the field of microscopic reconstructed Helmholtz energy can be obtained and, in turn, the reconstructed macroscopic energy. Namely, the reconstructed macroscopic energy, $\bar{\Psi}$, can be written as follows:
\begin{equation}\label{EQ17}
    \bar{\Psi}= \mathcal{O}\left(\bar{\psi} \right) = \mathcal{O}\left( f^{\psi}(\bar{\boldsymbol{\xi}}) \right) = \mathcal{O}\left( f^{\psi}(\boldsymbol{\bar{U}}\boldsymbol{\bar{U}}^{*}\boldsymbol{\xi}) \right),
\end{equation}
where $\mathcal{O}$ is an appropriate upscaling operator, like the volume average, $\langle\bullet\rangle$, in the homogenization problem.
$\bar{\Psi}$ is therefore a function of the number of POD modes considered, $N_{POD}$, and can be exploited to define a reconstruction error on a normalized macroscopic energy:
\begin{equation}
    err_{\Psi}(t) = \frac{\Psi(t) - \bar{\Psi}(t)}{\Psi_{mean}},
\end{equation}
where $\Psi_{mean}$ is the mean value of the energy data in the dataset among all the possible values at any time. Thus, $err_{\Psi}$ is quantifying how high the reconstruction error is relative to the mean of the energy values in the dataset. It is remarked that the energy is defined with respect to a reference value, which, in this case, has been taken to 0 for the undeformed configuration. 

Based on the reconstruction error as a function of the number of retained POD modes, it is feasible to select an optimal number, $N_{POD}$, that ensures the reconstruction of the Helmholtz free energy at the macroscale is within a predefined threshold. Intriguingly and logically, this threshold for energy reconstruction is mirrored in the converging values of the loss function observed in the training learning curves of the energy network, $NN^{\Psi}$. This phenomenon is elucidated as follows: the internal state variables vector, $\boldsymbol{Z}$, serves as an input to the energy network. The extent of microscopic information it contains is directly proportional to the number of retained POD modes. Consequently, with a fixed number of POD modes, the macroscopic energy reconstruction error can be quantified, delineating the upper limit of what the energy network can learn. It is unrealistic to expect the network to assimilate more precise information than what is inherent in the data.

Based on the reconstruction error as a function of the number of retained POD modes, an optimal number, $N_{POD}$, can be selected to ensure the Helmholtz free energy reconstruction at the macroscale remains within a predefined threshold. This threshold aligns with the converging values of the loss function in the energy network, $NN^{\Psi}$, indicating the extent of microscopic information captured by the retained POD modes.

The advantage offered by this procedure is the a priori estimation of the attainable training error of the energy network. This allows for an informed decision on the number of POD modes to retain, ensuring efficient and effective training of the network.

When it is not possible to calculate the energy at the microscale, the benefit of a priori error estimation is lost. Nevertheless, the number of retained POD modes can be introduced as an additional hyperparameter to regulate the overall loss reduction. By adjusting the number of POD modes, the neural network can achieve a progressively smaller training error until it reaches an asymptotic value, at which point the benefit of increasing the retained POD modes is outweighed by the increased computational time due to the network's complexity. However, this procedure is beyond the scope of this work and is reserved for future research.

\subsection{Uniqueness of the energy functions}\label{S3.3}
The energy network undergoes a supervised training process in which data of energy, stress and rate of dissipation have to be provided. As already pointed out in \cite{Masi_ETANN}, \cite{he2022thermodynamically}, it is not strictly necessary to include the output of the energy in the trainable variables, since compliance with the first law is guaranteed by construction, regardless of the accuracy of the energy data reproduction. In the following, the possibility of eliminating the rate of dissipation data is also taken into account. 

A first consideration to make is on the uniqueness of the partition of the internal power in the first law. For isothermal processes, the first law is rewritten as follows: $\boldsymbol{\sigma}:\dot{\boldsymbol{\epsilon}}=\dot{\psi}+d$, in which it is seen that the internal power is partitioned into the rate of stored energy and the rate of dissipated energy. Collins and Houlsby \cite{collins1997application}, have remarked that the decomposition is in general not unique. To show this, we assume to define a constitutive model by means of the two functions Helmholtz free energy density $\psi_1=f^{\psi}_{1}(\boldsymbol{\epsilon}, \boldsymbol{z})$ and rate of dissipation  $d_1$. An additional model is considered, characterized by $\psi_2=f^{\psi}_{1}(\boldsymbol{\epsilon}, \boldsymbol{z})+f^{\psi}_{2}(\boldsymbol{z})$ and $d_2$. Substituting the functions defining the two models within the first law, the following expressions are found:
\begin{equation}\label{EQ18}
    \begin{split}
        &\dot{\psi}_1=\frac{\partial f^{\psi}_1}{\partial \boldsymbol{\epsilon}}:\dot{\boldsymbol{\epsilon}}+\frac{\partial f^{\psi}_1}{\partial \boldsymbol{z}}\cdot\dot{\boldsymbol{z}} = \boldsymbol{\sigma}:\dot{\boldsymbol{\epsilon}} - d_1, \\
        &\dot{\psi}_2=\frac{\partial f^{\psi}_1}{\partial \boldsymbol{\epsilon}}:\dot{\boldsymbol{\epsilon}}+\frac{\partial f^{\psi}_1}{\partial \boldsymbol{z}}\cdot\dot{\boldsymbol{z}}+\frac{\partial f^{\psi}_2}{\partial \boldsymbol{z}}\cdot\dot{\boldsymbol{z}} = \boldsymbol{\sigma}:\dot{\boldsymbol{\epsilon}} - d_2.
    \end{split}
\end{equation}
It is straightforward to see that the stress, which is the quantity of interest, is identical between the two models, $\boldsymbol{\sigma}=\partial f^{\psi}_1/\partial\boldsymbol{\epsilon}=\partial f^{\psi}_2/\partial\boldsymbol{\epsilon}$, even though the free energy model is different, as well as the rate of mechanical dissipation, that results to be $d_2=d_{1}-\frac{\partial f^{\psi}_2}{\partial \boldsymbol{z}}\cdot\dot{\boldsymbol{z}}$. 

Starting from this consideration, using both energy and dissipation rate data obtained from specific energy models could impose an unnecessary requirement during the training of the energy network. TANN can determine an adequate energy function based solely on stress data. It is important to recall that the aim of TANN is to achieve thermodynamically admissible constitutive predictions, meaning consistent predictions of stress and state variable increments based on strain increments. Excluding the dissipation term from the training loss, specifically $\lVert d+\frac{NN^{\Psi}}{\partial\boldsymbol{Z}}\dot{\boldsymbol{Z}} \rVert$, would prevent the network from learning the exact rate of the dissipation model included in the data. However, it would still ensure the thermodynamic consistency of the predictions and the correct reproduction of the stress tensor. 

A further benefit of excluding the energy terms in the training data is the enhanced applicability of TANN for data derived from laboratory experiments, where measuring energy quantities is often challenging. 

Therefore, the following loss function will be considered:
\begin{equation}\label{EQ19}
    \mathcal{L} = w^{\boldsymbol{\sigma}}\ell^{\boldsymbol{\sigma}} +  w^{\{d\}}\ell^{\{d\}}.
\end{equation}

\section{Applications}\label{S4}
This section explores three applications of the proposed approach. Sections \ref{S4.1} and \ref{S4.2} focus on learning the homogenized behavior of periodic unit cells with inelastic microstructures, while Section \ref{S4.3} addresses the data-driven derivation of a macroelement for the horizontal response of a monopile embedded in a clay layer.

In these examples, both the macroscopic stress-strain (or force-displacement) response and the reconstruction of microscopic fields are examined, with particular emphasis on the influence of the number of retained POD modes for describing the components of the ISV vector. Accurately reconstructing microscopic fields, especially in systems using periodic Representative Unit Cells (RUCs), requires the fulfillment of boundary conditions. Although the POD-TANN approach effectively reduces system dimensionality and trains efficient models, it does not inherently enforce periodic boundary conditions during reconstruction, which may introduce errors. These errors depend on the quantity of interest and the various error sources involved, particularly in strain field reconstruction, where errors in fulfilling periodic BCs are typically related to the number of retained POD modes. 

\subsection{RUC with ellipsoidal inclusion}\label{S4.1}
The first example focuses on an RUC with an ellipsoidal inclusion. Both the matrix and inclusion follow an elasto-plastic model with isotropic hardening and the Drucker-Prager yield criterion. The material parameters are provided in Table \ref{TAB2}.

\subsubsection{Data generation}\label{S4.1.1}
The training dataset was generated by applying random strain increments to the RUC, starting from an initial compression volumetric strain of -5e-4.  This approach was chosen to avoid generating data in the positive (tensile) volumetric domain, which is rarely relevant for geomechanics applications. The strain paths were created by sampling increments from a standard normal distribution with a mean of 0.0 and a standard deviation of 5e-4. To ensure that a significant portion of the stress points fall within the plastic regime, a limit of 0.015 was imposed on the second deviatoric strain invariant.

Strain paths consisting of 1000 increments were considered. The initial dataset was composed of five random strain paths. The final dataset was then obtained through data augmentation. This involved rotating the second-order tensors of the microscopic stress and strain fields using a randomly sampled 3D rotation matrix, $\boldsymbol{R}$, while preserving invariant quantities such as energy. This augmentation not only reduced computational time for dataset generation but also allowed the neural network to learn objectivity from the data.

\begin{table}
    \centering
    \begin{tabular}{ |c|c|c|c|c|c|c| }
        \hline
        Material & $E$ (kPa) & $\nu$ (-) & $\phi$ (°) & $\tilde{\psi}$ (°) & $c$ (kPa) & $H$ (kPa) \\
        \hline
         Matrix & 5500 & 0.3 & 32 & 32 & 10 & 4000 \\ 
         Inclusion & 6500 & 0.3 & 30 & 30 & 12 & 3500 \\ 
        \hline
    \end{tabular}
    \caption{Constitutive parameters used for the RUC with ellipsoidal inclusion. $E$ the Young’s modulus, $\nu$ the Poisson’s ratio, $\phi$ the friction angle, $\Tilde{\psi}$ the dilatancy angle, $c$ the effective cohesion and $H$ the hardening modulus.}
    \label{TAB2}
\end{table}

The ICs are obtained by collecting the elastic and plastic microscopic strains, along with the maximum second invariant of the plastic strains at the Gauss points of the computational model at each increment, expressed as $\boldsymbol{\xi}=\left[\boldsymbol{\epsilon}^{el}_i, \boldsymbol{\epsilon}^{pl}_i, J_{2, max}^{pl}\right]$ at all Gauss points. Since the hardening in this example is isotropic and linear, and there is no softening, $J_{2, max}^{pl}$ serves as the hardening variable (see \cite{abaqus_manual}). These microscopic state variables are sufficient to fully describe the microscopic state at each Gauss point of the RUC.

\begin{figure}
\centering
    \includegraphics[width=1.0
\textwidth]{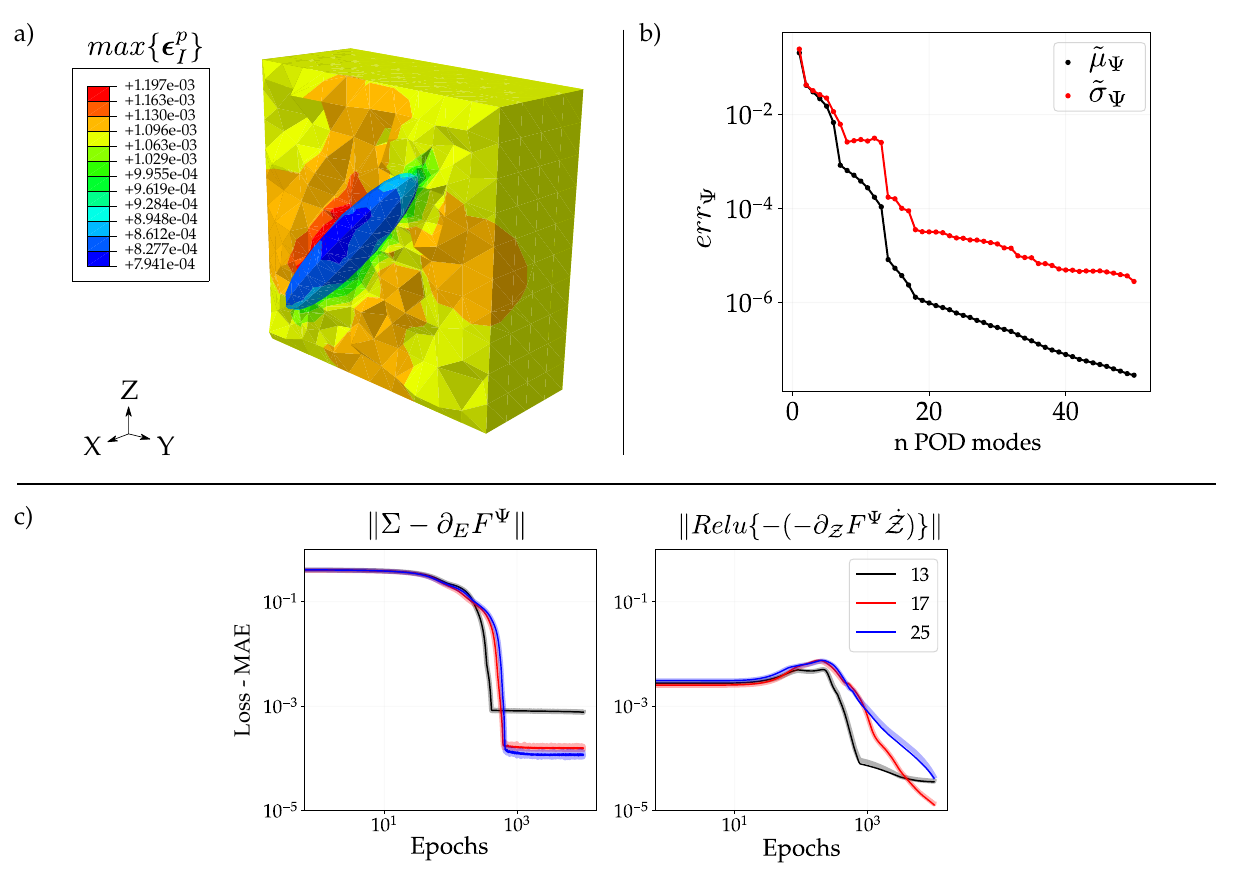}
\caption{a) Computational model of the RUC, representation of the field of maximum principal microscopic plastic strains. b) Normalized reconstruction error $(\Psi - \Psi)/\Psi_{max}$ of the macroscopic energy based on considered number of POD modes. $\tilde{\mu}_{\Psi}$ and $\tilde{\sigma}_{\Psi}$ are the mean and standard deviation of $err_{\Psi}$ over all the data samples in the training set; $\Psi_{max}$ is the maximum energy value in the training set. c) Loss curves the macroscopic stress and the regularization term on the rate of dissipation, considering 13, 17 and 25 POD modes.}
\label{FIG5}
\end{figure}

With 10675 linear tetrahedral elements, the computational model is computationally expensive. A total of 138775 degrees of freedom have been collected in the IC matrix. As shown in Figure \ref{FIG5}-b, with a number of POD coefficients greater than 13, the energy reconstruction error, as defined in section,  is reduced to 1e-4.

The reduction in dimensionality of the ICs prior to training the energy network has been quantified using a compression ratio, $CR$, defined as follows:
\begin{equation}\label{EQ20}
    CR=(1-dim(\boldsymbol{\mathcal{Z}})/dim(\boldsymbol{\xi}))\quad \%
\end{equation}

With 25 POD modes, the compression is $CR=99.98\%$. This substantial reduction is achievable due to the redundancy in the information contained within the set of internal coordinates. Although the material's heterogeneity introduces variability in the strain fields, the gradients are primarily concentrated around the inclusion. Many Gauss points within the matrix, especially those far from the inclusion, exhibit strain field values that change marginally relative to one another, enabling such high compression.

\subsubsection{Training of the TANN}\label{S4.1.2}
To train the Helmholtz free energy network a supervised learning procedure has been adopted, collecting inputs and outputs as detailed in Section \ref{S3}. All the inputs and outputs have been normalized to range between -1 and 1, with the exception of the Helmholtz energy and the rate of mechanical dissipation, for which positive and normalized values between 0 and 1 have been used. As schematically depicted in Figure \ref{FIG1}, the input of the network has variable dimensionality, depending on the number of retained POD modes, in agreement with the following expression: $6 + 2\cdot N_{POD}$. Despite the latter variability, the training of the TANN has been performed keeping constant the number of hidden layers and neurons of the energy network, as detailed below.

The energy network is a feed forward net with a single hidden layer of 100 neurons with quadratic activation function of the form $p^{(k)}_j = \mathcal{A}^{k}(r^{(k)}_j)=r^{(k) 2}_j$, and a linear single output layer with biases set to zero. The latter guarantees the reference energy to be zero at an initial zero-valued state. The Nadam optimizer has been adopted, see \cite{dozat2016incorporating}, with a learning rate of 5e-5. Weight and biases have been initialized using the Glorot uniform initializer, see \cite{glorot2010understanding}. A constant mini-batch size of 1000 has been used.

\begin{figure}
\centering
    \includegraphics[width=1.0
\textwidth]{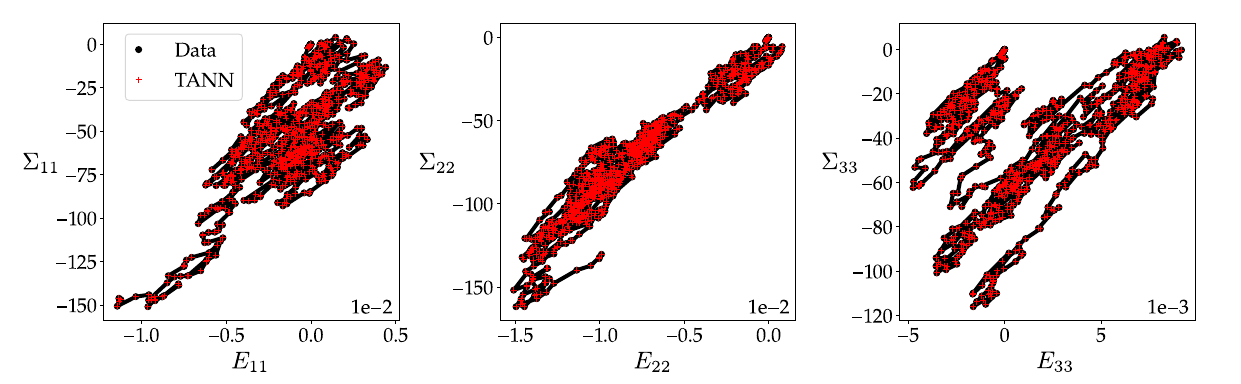}
\caption{Prediction in inference mode of the TANN's energy network trained considering 25 POD modes of the RUC with the ellipsoidal inclusion. The RUC is subjected to an unseen random 3D strain path. Stress components are reported as a functions of the conjugate strain components.}
\label{FIG10}
\end{figure}

\subsubsection{Inference of the TANN}\label{S4.1.3}
The capabilities of the trained network are tested for an unseen random 3D strain path, as reported in Figure \ref{FIG10}. The computed Mean Absolute Error (MAE) is of the order of 1e-4. 
\begin{figure}
\centering
    \includegraphics[width=1.0
\textwidth]{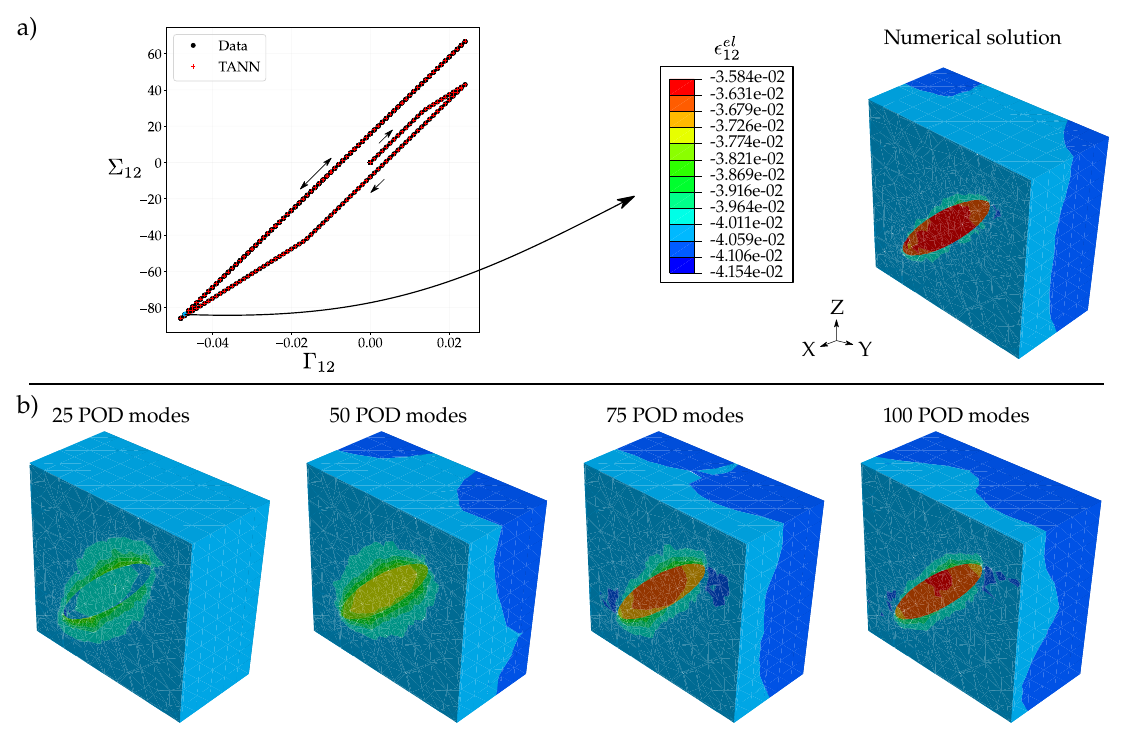}
\caption{a) Prediction in inference mode of the TANN's energy network trained considering 25 POD modes. On the right there is the microscopic elastic strain field corresponding to a point in the first reloading branch of the path. b) Reconstructed elastic strain field considering 25, 50, 75 and 100 POD modes.}
\label{FIG6}
\end{figure}
Figure \ref{FIG6} illustrates the application of the TANN in inference mode along an unseen pure-shear cyclic strain-driven path, well beyond the training range. The TANN used for predictions is trained with 25 POD modes. The figure demonstrates the network's ability to account for the evolution of internal hidden variables, successfully predicting the expansion of the elastic domain induced by isotropic hardening. Additionally, the model accurately predicts the permanent effect of hardening. After a complete stress reversal in one cycle, the expansion of the elastic domain prevents re-entering the elastic-plastic regime during subsequent cycles of the same amplitude. Consequently, the neural network predicts an elastic response along the load-unload line to the left of the origin, aligning with the exact solution and the physics of the problem.

The use of POD enables the fast reconstruction of the detailed fields of the ICs. Figure \ref{FIG6}-b depicts the reconstructed elastic deformation field using different numbers of POD modes. The reconstruction of the field with 25 modes achieves a MAE of approximately 1e-3, which is satisfactory for this application. An even more accurate reconstruction of the microscopic fields can be obtained by increasing the number of POD modes; for instance, using 100 modes reduces the reconstruction MAE to 1e-5.

The macroscopic constitutive behavior, quantified in terms of stresses, is well reproduced (with an MAE of about 1e-4), despite the less accurate reconstruction of the microscopic fields (1e-3). This is naturally induced by the averaging process used to upscale the microscopic response. Therefore, depending on the application, one can choose the appropriate number of POD modes for the TANN to reconstruct the IC's fields with the required accuracy. The selection of the number of modes can be made after performing the singular value decomposition (SVD), as described in Section \ref{S3.2}. The decomposition needs to be done only once, while only the energy network must be re-trained each time the selected number of POD modes changes. This is a clear advantage of our approach compared to previous ones using autoencoders (\cite{Masi_2022}, \cite{Masi_ETANN}). Moreover, as noted in Section \ref{S3.1}, autoencoders cannot provide hierarchically sorted reduced data. This implies the need to train them together with the TANN's energy network, necessitating a new, coupled training process each time the dimensionality of the reduced space is changed, which is computationally expensive.

\subsection{RUC with leaf-shaped inclusion}\label{S4.2}
The second example features a cubic representative unit cell (RUC) with a leaf-shaped inclusion, included to address a more complex inclusion topology. While numerous studies in the literature focus on ellipsoidal particle inclusions, such as \cite{jones2022neural}, fewer explore generic geometries like the one presented in this section. This example demonstrates the versatility of the method when applied to inclusions of arbitrary shapes.

\subsubsection{Data generation}\label{S4.2.1}
The constitutive relationships for both the matrix and the inclusion are elasto-plastic, with linear isotropic hardening and the Drucker-Prager yield criterion. The constitutive parameters are detailed in Table \ref{TAB3}. The internal configurations (ICs) used are the field variables sufficient to microscopically describe the state of the material, including the elastic and plastic strains and the hardening variables at the Gauss points of the finite element (FE) model. This setup results in a total of 92,703 IC degrees of freedom. The neural network is trained similarly to the previous case, with dataset augmentation by rotation. The training dataset starts with 5 sets of 1000 samples, each randomly rotated 5 times, and an initial compression volumetric strain threshold of -5e-4 is set for data generation.

\begin{table}[]
    \centering
    \begin{tabular}{ |c|c|c|c|c|c|c| }
        \hline
        Material & $E$ (kPa) & $\nu$ (-) & $\phi$ (°) & $\tilde{\psi}$ (°) & $c$ (kPa) & $H$ (kPa) \\
        \hline
         Matrix & 5500 & 0.3 & 25.4 & 25.4 & 5 & 3000 \\ 
         Inclusion & 6500 & 0.3 & 30 & 30 & 12 & 2500 \\ 
    \hline
    \end{tabular}
    \caption{Constitutive parameters used for the RUC with leaf-shaped inclusion. Symbols are defined in Table \ref{TAB2}.}
    \label{TAB3}
\end{table}

\subsubsection{Training of the TANN}\label{S4.2.2}
\begin{figure}
\centering
    \includegraphics[width=1.0
\textwidth]{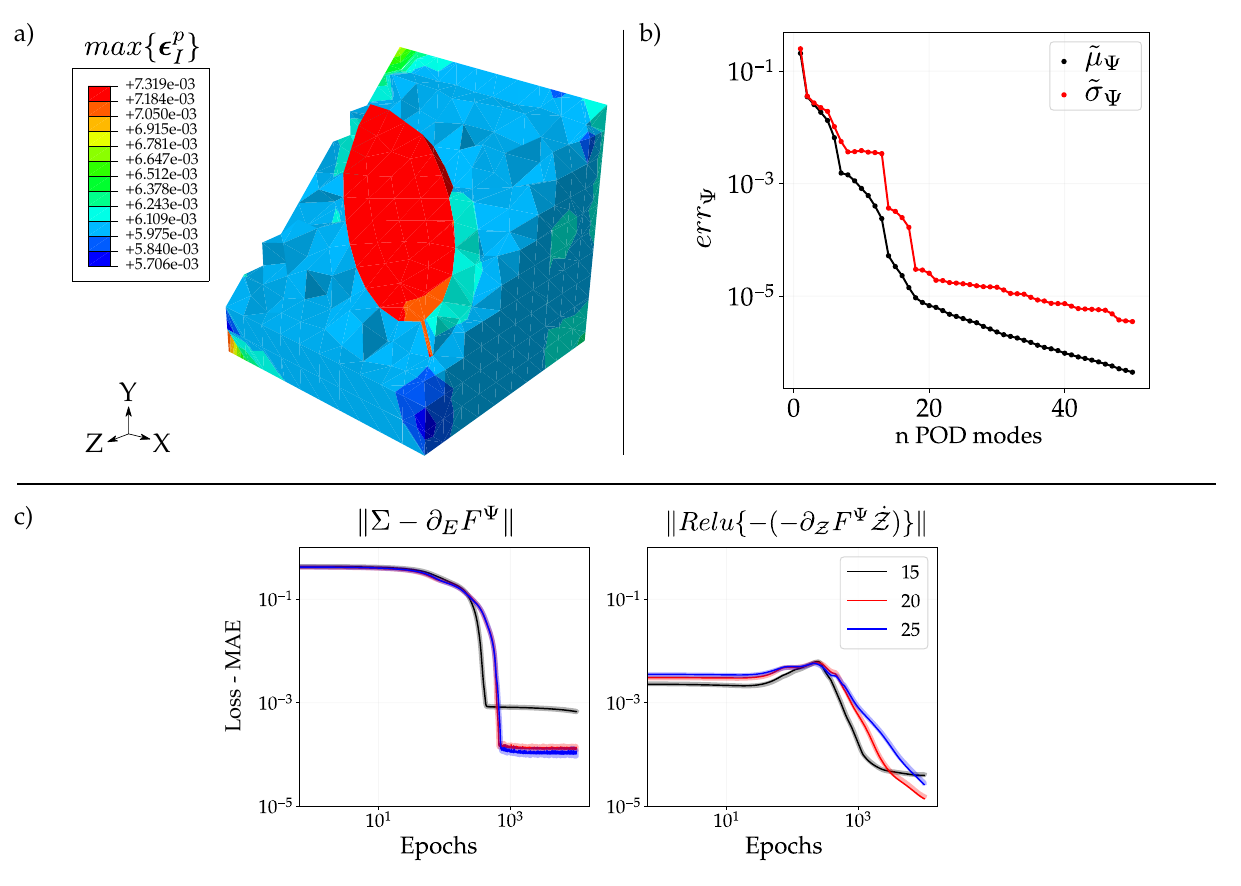}
\caption{a) Computational model of the RUC, representation of the field of maximum principal microscopic plastic strain. b) Normalized reconstruction error $(\Psi - \Psi)/\Psi_{max}$ of the macroscopic energy based on considered number of POD modes. $\tilde{\mu}_{\Psi}$ and $\tilde{\sigma}_{\Psi}$ are the mean and standard deviation of $err_{\Psi}$ over all the data samples in the training set; $\Psi_{max}$ is the maximum energy value in the training set. c) Loss curves of the macroscopic stress and the regularization term for imposition of the second Law, considering 15, 20 and 25 POD modes.}
\label{FIG7}
\end{figure}

The training of the TANN has been achieved using the same setting detailed in section \ref{S4.1.2}. Figure \ref{FIG7}-c presents the learning curves obtained during the training process. Remarkably, the neural network training converges after only 1000 epochs. The figure displays the learning curves for 15, 20, and 25 POD modes. A significant improvement is observed from 15 to 20 modes, with the plateau value decreasing by almost one order of magnitude. This improvement is also evident in Figure \ref{FIG7}-b, where the average energy reconstruction error and its standard deviation decrease by approximately one order of magnitude. Beyond 20 POD modes, the error curve's slope remains nearly constant as the number of modes increases. This is reflected in the minimal difference between the stress loss curves for 20 and 25 modes, although a further decrease in the plateau value is noted with 25 modes. With 25 modes, the compression ratio is $CR=99.97\%$.

\subsubsection{Inference of the TANN}\label{S4.2.3}
\begin{figure}
\centering
    \includegraphics[width=1.0
\textwidth]{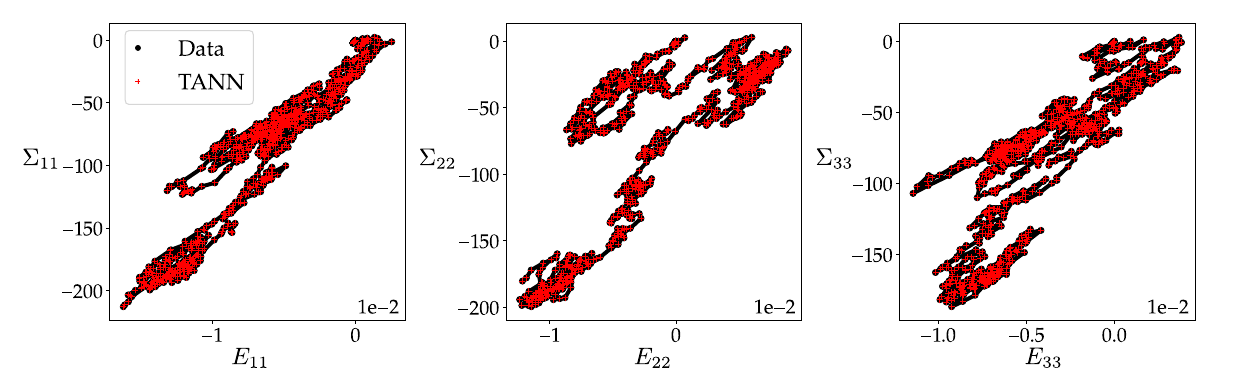}
\caption{Example of use of the trained energy network for the RUC with leaf-shaped inclusion in inference mode. The RUC is subjected to a 3D strain driven unseen random strain path. Stress VS conjugate strain components.}
\label{FIG11}
\end{figure}

Similar to the previous cases, an example of the use of the trained network in inference mode is reported in Figure \ref{FIG11}, to show the good reproduction of an unseen 3D random strain path. 

\begin{figure}
\centering
    \includegraphics[width=1.0
\textwidth]{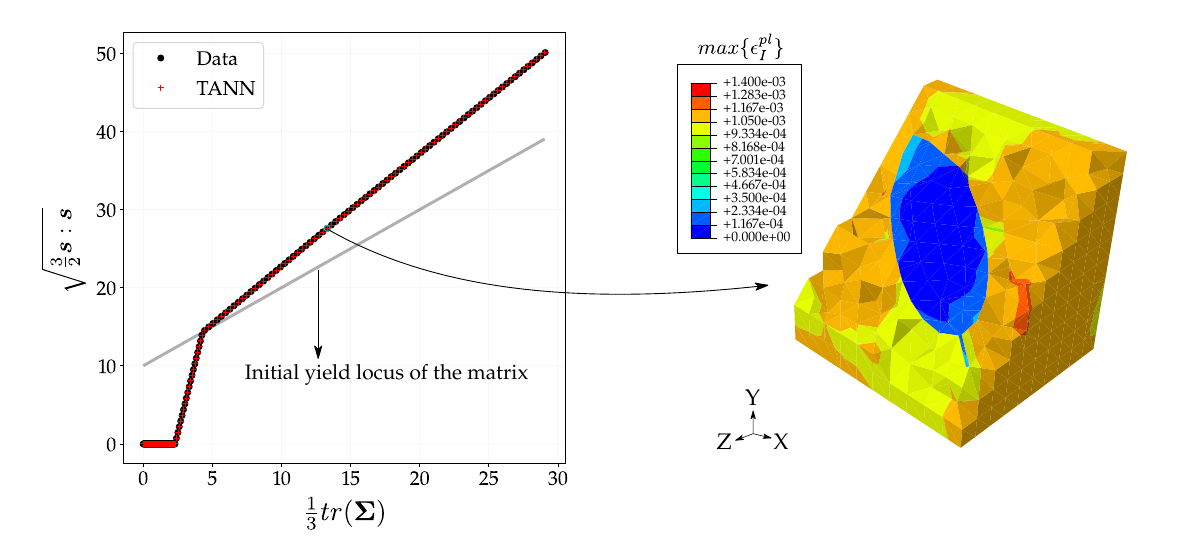}
\caption{Example of use of the trained energy network for the RUC with leaf-shaped inclusion in inference mode. The reproduced path is a triaxial strain-driven compression test, starting from an isotropic compression.}
\label{FIG8}
\end{figure}

Figure \ref{FIG8} illustrates another example of the use of the trained TANN's energy network in inference mode. This case reproduces a strain-driven compression triaxial test. The RUC is initially subjected to volumetric deformation, followed by simultaneous increases in the strain components $E_{zz}$, $E_{xx}$ and $E_{yy}$ at a constant ratio, allowing for both volumetric and shear deformation. At the beginning of the shear phase, the stress ratio is uniquely determined by the homogenized elastic volumetric and shear stiffnesses, up to the yielding point. The first yield surface, appearing as a line in the invariant stress plane, is associated with the matrix's constitutive behavior, as shown in grey in Figure \ref{FIG8}. The matrix elements are the first to yield. Due to stress redistribution, additional Gauss points enter the elasto-plastic regime, altering the ratio between the corresponding volumetric (still elastic) and elasto-plastic deviatoric tangent stiffnesses. This also applies to the inclusion, whose elements enter the elasto-plastic regime due to stress concentration at its boundary with the matrix. Once the matrix yields, several elements within the inclusion plastify, although its core remains elastic under the applied strain amplitude.

\begin{figure}
\centering
    \includegraphics[width=1.0
\textwidth]{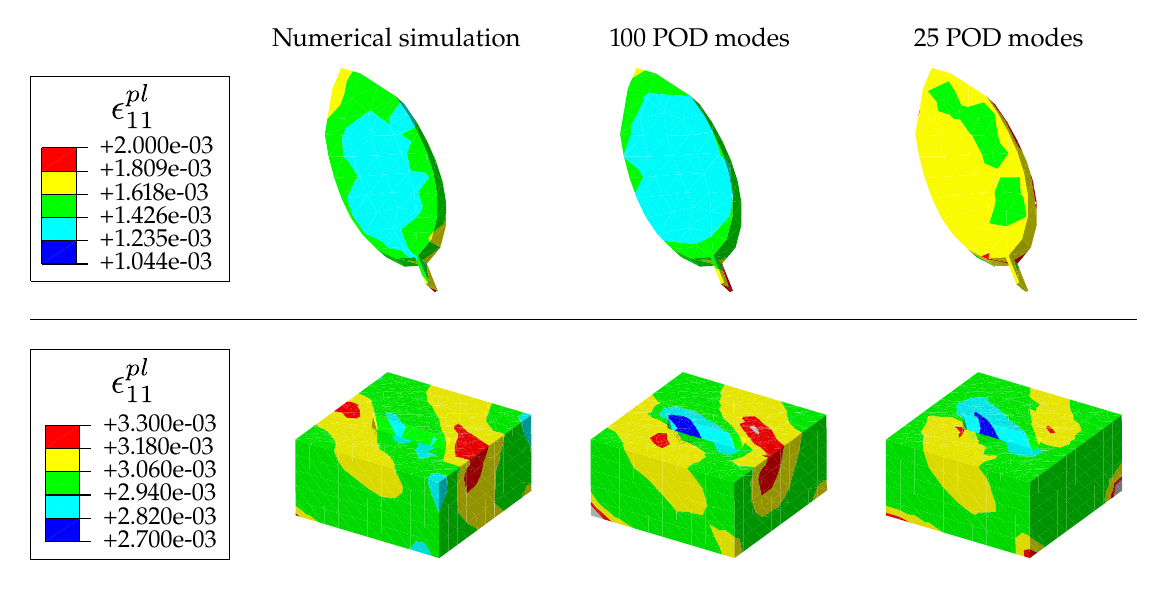}
\caption{Reconstructed plastic strain field ($\epsilon^{pl}_{11}$ component) considering 25 and 100 POD modes. The numerical results refer to the last increment of the triaxial strain path shown in Figure \ref{FIG8}. }
\label{FIG9}
\end{figure}

Figure \ref{FIG9} shows the microscopic plastic strain field for the first normal component at the final increment of the strain path depicted in Figure \ref{FIG8}. The POD-reconstructed fields are compared to numerical simulation results. As anticipated, increasing the number of POD modes improves the agreement. Although the plastic field of the leaf-shaped inclusion is replicated with less accuracy, 25 modes are sufficient to capture the pattern of the major plastified zones in the matrix. The mean absolute reconstruction error is approximately 5e-4. This error reduces to 5e-5 when 100 POD modes are considered.

\subsection{Macroelement of a monopile in saturated clay}\label{S4.3}
The third example explores the possibility of utilizing the proposed POD-TANN framework to learn the energy potential of a macroelement. The focus is on a monopile foundation, which is the most common solution for offshore wind turbines installations in shallow waters.

Traditionally, the evaluation of the response of these types of foundations has been based on simplified approaches, such as those of p-y curves, as in \cite{matlock1970correlation} and \cite{mayoral2005determination}, and more rarely on complex numerical analyses of the single pile and the foundation soil. Alternatives to these approaches are macroelements. These macroelements maintain the simplicity of use found in p-y curves, but they allow for the direct consideration of complex aspects of soil-structure interaction on the macroscopic response of the system, such as the progressive degradation of the soil's stiffness and resistance, friction between the pile and the ground, and the hysteretic damping response of the soil, among others.

In this section, a macroelement capable of replicating the horizontal undrained cyclic response in terms of force-displacement of a monopile in a saturated clay layer is presented. The macroelement results from training a TANN on a numerical database obtained from finite element simulations of the macroscopic system.

The proposed methodology utilizes the energy formulation in terms of the dual energy potential of the Helmholtz free energy, namely the Gibbs free energy. Engineering structures are often subjected to known forces, and the displacements induced by external solicitations are the unknowns. Thus, it is convenient to formulate the problem in terms of the Gibbs potential and, by its partial differentiation, obtain the horizontal displacements induced by the resultant of the external forces acting on the pile.

The TANN is combined with Proper Orthogonal Decomposition and applied to a macroscopic geotechnical system. Eventually, the studied 3D system is condensed into a macroelement with one degree of freedom. 

The macroscopic state of the pile-soil system is denoted by $\mathcal{S}=\left[F_H, \mathcal{\boldsymbol{Z}}\right]$, where $F_H$ represents the resultant of the horizontal forces that are active at the pile head, and $\boldsymbol{\mathcal{Z}}$ the vector of internal state variables of the macroscopic system, which are determined based on the IC, $\boldsymbol{\xi}$. Figure \ref{APP:macroelement:neural_network} illustrates a schematic representation of the procedure presented below. 
\begin{figure}
\centering
    \includegraphics[width=1.0
\textwidth]{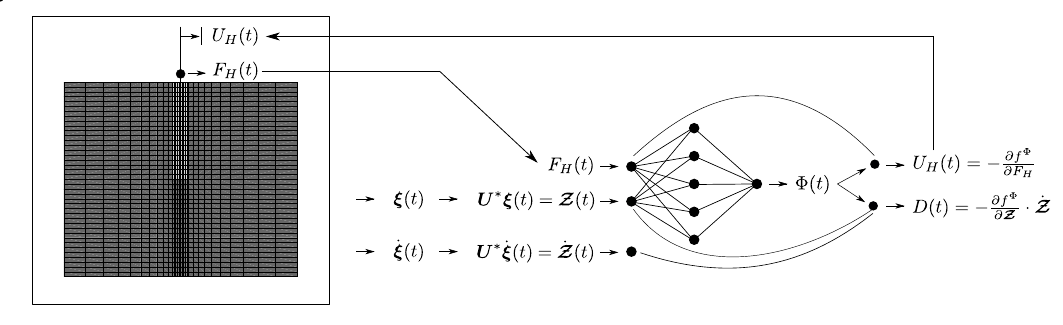}
\caption{Schematic of the problem setting. The data obtained from the numerical FE analysis of the geotechnical system are fed into the TANN to obtain the macroscopic response of the system in terms of horizontal force-displacement.}
\label{APP:macroelement:neural_network}
\end{figure}

In order to learn the Gibbs free energy potential function of the macroscopic system, a neural network denoted by $f^{\Phi}$ is trained.
The network receives as input the macroscopic state of the system and returns as output the Gibbs free energy, $\Phi$. 
By partial derivation of $f^{\Phi}$, two further quantities are obtained: (i) the horizontal displacement, $U_H = -\frac{\partial f^{\Phi}}{\partial F_H}$, and the rate of mechanical dissipation, $D=-\frac{\partial f^{\Phi}}{\partial \mathcal{\boldsymbol{Z}}}\dot{\boldsymbol{\mathcal{Z}}}$.

\subsubsection{Computational model}\label{S4.3.1}
The 3D finite element model has been set up in ABAQUS \cite{abaqus_manual}. To simulate the undrained response, the soil is modeled as a single-phase medium with a saturated density of $\rho_{SAT} = 2 \ \text{Mg/m}^3$, and the analyses are conducted in terms of total stresses.

An elastic-plastic constitutive model with kinematic hardening and the Von-Mises yield criterion is used for the soil. The undrained resistance is a linear function of the vertical coordinate as follows: $S_u = 10 + 0.3p'(z)$, in $\text{kPa}$, with $p'$ the effective mean stress. The elastic law of the soil is linear isotropic, with the Young's modulus values depending on $z$, as follows: $E(z) = E_{ref}\left( \frac{p'(z)}{p_{ref}'}\right)^{0.5}$, with $p_{ref}' = 30\text{kPa}$, $E_{ref} = 500 \ S_{u,ref}$, $S_{u,ref} = 20\text{kPa}$. The Poisson's ratio is $0.48$, to account for the potential presence of gas in the saturated porous medium, thereby reducing the volumetric stiffness of the equivalent single-phase (similar to what is done for gassy soil, see \cite{yang2020thermodynamic}).

The steel pile with a hollow circular section is modeled as isotropic linear elastic with the following parameters: $E = 210\text{GPa}$, $\nu = 0.3$, $\rho = 7.85\text{Mg/m}^3$.

The soil inside the pile cavity is modeled as elastic and not elasto-plastic, with the same elastic parameters already reported. This modeling assumption is adopted following an elasto-plastic analysis, according to which it has been found the plastic deformations of the soil inside the pile to be negligible.

\begin{figure}
\centering
    \includegraphics[width=1.0
\textwidth]{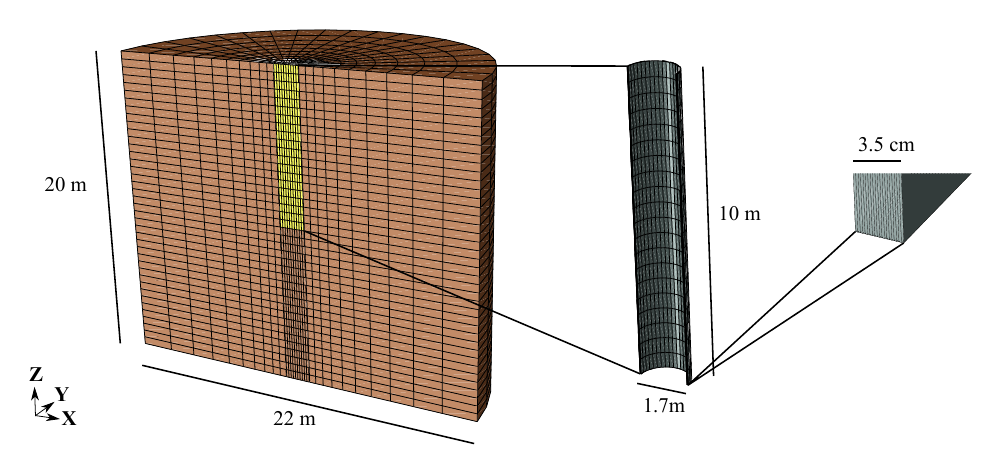}
\caption{Computational model and generated mesh of the soil-pile system.}
\label{APP:macroelement:mesh}
\end{figure}

A Coulomb frictional contact is modeled between the pile and the soil, with a coefficient of 0.5. Figure \ref{APP:macroelement:mesh} shows the dimensions of the computational model and the representation of the generated mesh.

\subsubsection{Data generation and training of the TANN}\label{S4.3.2}
\begin{figure}
\centering
    \includegraphics[width=1.0
\textwidth]{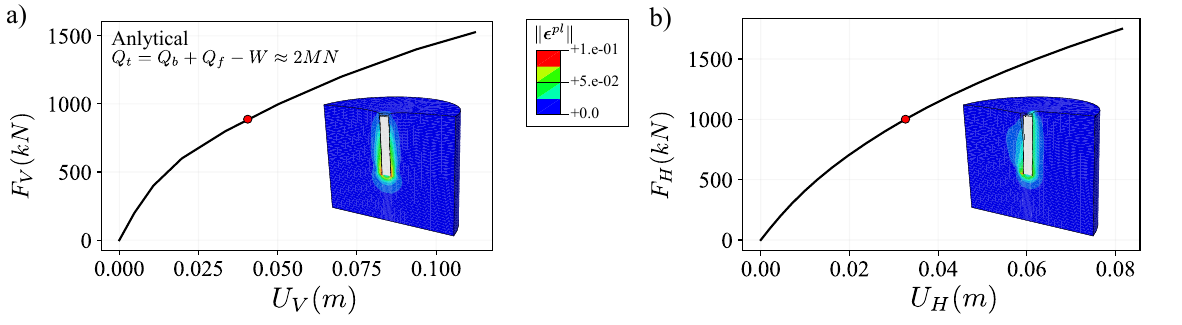}
\caption{Monotonic load curves for vertical (a) and horizontal (b) bearing capacity.}
\label{APP:macroelement:bearing}
\end{figure}

The TANN training database has been obtained by subjecting the pile to a vertical operating force equal to $800 \ kN$ and to random horizontal forces variable in sign and amplitude sampled from a uniform distribution with a maximum amplitude equal to $1 \ MN$, as highlighted by the red dots in Figure \ref{APP:macroelement:bearing}.

The soil-pile system is reduced to a macroelement with one degree of freedom, considering the horizontal displacement of the point at the pile head positioned on the pile axis, $U_H$, as a function of the resultant of the horizontal forces at the pile head, $F_H$. 

\begin{figure}
\centering
    \includegraphics[width=1.0
\textwidth]{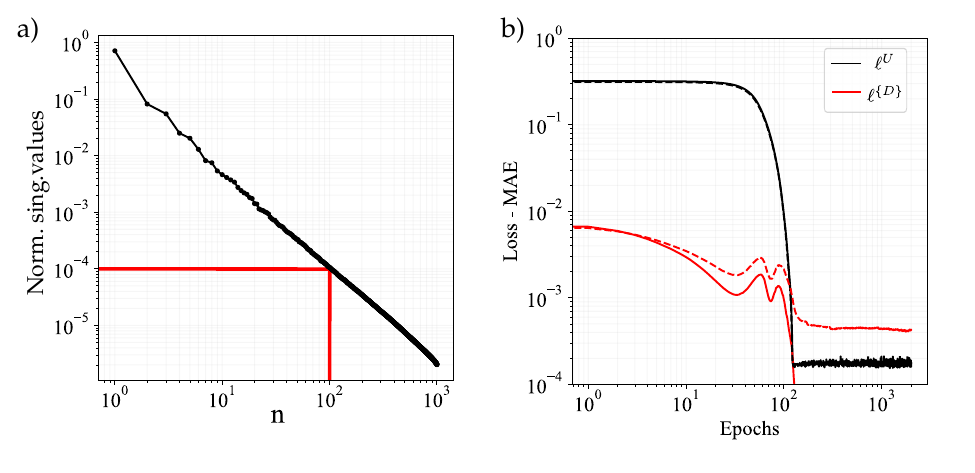}
\caption{a) Normalized singular values obtained via POD from the set of ICs. b) Training curves of the TANN. The solid lines refer to the training set while the dotted lines refer to the validation set, used to avoid over-fitting phenomena.}
\label{APP:macroelement:POD_results}
\end{figure}

The elastic and plastic strains, along with the hardening variable at each Gauss point of the FEM model, have been included in the IC set, onto which POD has been performed to obtain the ISVs. Figure \ref{APP:macroelement:POD_results}-a displays the normalized singular values obtained. The number of POD modes retained was determined based on the relative magnitude of these normalized singular values. A threshold value of 1e-4 was set, leading to the selection of 100 POD modes. This threshold influences directly the asymptotic value of the training losses of the TANN, as shown in Figure \ref{APP:macroelement:POD_results}-b, where the training stopped upon reaching 1e-4.

As already shown, the threshold of 1e-4 allows to achieve satisfactory results both in terms of accuracy in the reproduced response of the 1D equivalent system and of the reconstruction error of the microscopic fields. 

\subsubsection{Inference of the TANN}\label{S4.3.3}
For this kind of applications the possibility of reconstructing the fields used as internal coordinates is particularly appealing. One of the major drawbacks of the classical macroelement formulations is the impossibility of retrieving microscopic information, once the macroscopic behavior has been simulated. This poses clear difficulties in the engineering design, since often structural components are encapsulated by the macro-material whose response is of interest. Thus, an additional model is needed to link the global response of the classical macroelement to that of the structural components. As shown above, this in not needed with the proposed POD-TANN approach.

\begin{figure}
\centering
    \includegraphics[width=0.95
\textwidth]{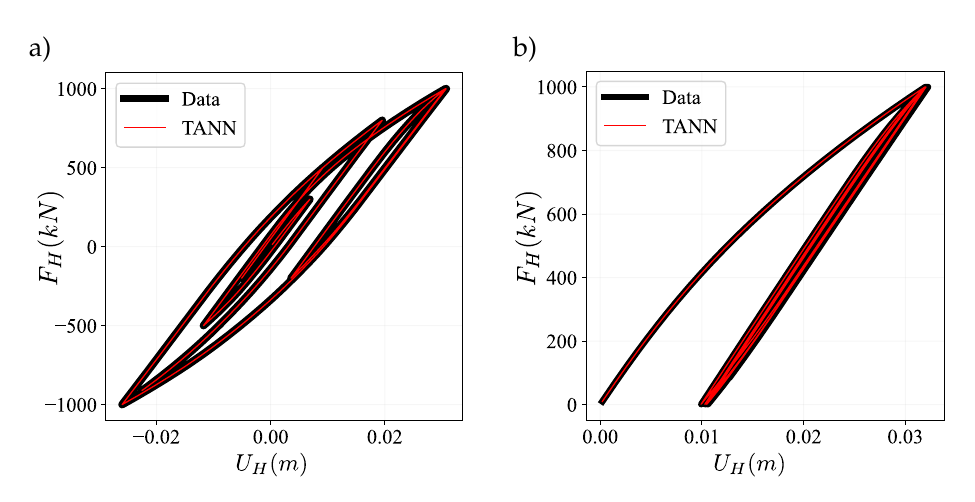}
\caption{TANN predictions in inference of a strain path outside the training set for cycles of positive and negative amplitude (a) and asymmetric cycles (b).}
\label{APP:macroelement:Results}
\end{figure}

Figure \ref{APP:macroelement:Results} shows the inference responses of the trained TANN. In particular, in Figure \ref{APP:macroelement:Results}-a the amplitude of the stress undergoes a total inversion between -1 and +1 MN. In Figure \ref{APP:macroelement:Results}-b, an asymmetric, positive cycle is shown. The relative MAE is of the order of 1e-4 and the displacements induced by the loading cycles are well reproduced. 

\begin{figure}
\centering
    \includegraphics[width=0.95
\textwidth]{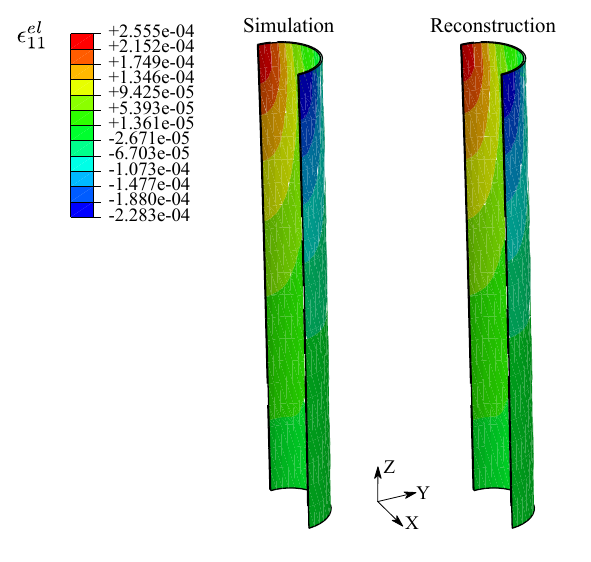}
\caption{Comparison between the deformation field $\boldsymbol{\epsilon}_{11}^{el}$ of the pile obtained from numerical simulations and reconstructed using POD.}
\label{APP:macroelement:Results_reconstruction}
\end{figure}

The potential of the method is particularly evident for the reconstruction task. Figure \ref{APP:macroelement:Results_reconstruction} shows the elastic deformation field, $\boldsymbol{\epsilon}^{el}$ of the pile simulated and reconstructed with the use of the POD, at the end of the last cycle of the load path of Figure \ref{APP:macroelement:Results}-b. 

The reconstruction error, expressed in MAE, is of the order of 1e-6. It is remarked here that the threshold of the normalized singular values is 1e-4, but this does not mean that the MAE of the reconstruction error has to be of the same order. The reconstruction error is computed as $\lVert \boldsymbol{\xi} - \tilde{\boldsymbol{U}}\boldsymbol{\mathcal{Z}} \rVert$, therefore it decreases as a function of the considered POD modes, but not necessarily with the same rate.  

By exploiting the known constitutive relationship, the stress field can be reconstructed with high accuracy and consequently the internal actions can be calculated for structural design.

\section{Discussion and perspectives}\label{S5}

\subsection{Applicability to complex inelastic systems}\label{S5.1}
The effectiveness of the proposed methodology is demonstrated through three applications of increasing complexity. These include the homogenization of two inelastic RUCs with continuous microstructures, featuring inclusions of increasing complexity. Additionally, a complex three-dimensional geotechnical model of a monopile embedded in a clay layer and subjected to horizontal loading is analyzed to showcase the procedure's applicability in deriving macroelements.

At the core of the methodology is the use of POD to hierarchically select the components of the ISV vector from the set of ICs. Despite the linear nature of this dimensionality reduction, the resulting compression ratios in the proposed examples - function of the ratio between the dimensionality of the ISV vector and that of the IC set - are highly satisfactory, consistently exceeding 90\%. In the RUC examples, IC sets comprising approximately 100000 degrees of freedom are reduced to an ISV vector with only 25 components, yielding a $CR$ of 99.97\%. This high compression ratio significantly reduces the need for complex hidden layers of the TANN energy network. In the examples considered, a network with a single hidden layer of 100 neurons proved sufficient, and in most cases, training converged rapidly, typically after 1000 epochs of optimization. The entire procedure requires roughly five minutes of runtime on standard 64-bit machines with 4 cores. Thus, the method is not only accurate, with a mean absolute error of around 1e-4 in macroscopic stress predictions, but also highly efficient.

The proposed POD-TANN approach also may serve as a valuable tool for materials with more complex microstructures. This is particularly relevant for materials exhibiting strong heterogeneities and nonlinearities. To further test the potential of this approach in the presence of strong heterogeneities, an additional RUC is investigated in Appendix B, featuring continuous spatially correlated random fields of constitutive parameters. In this example we show again that number of the significant POD modes remains low and comparable with the examples presented in Sections 4.1 and 4.2. Morevoer, we report high accuracy under high compression rates.

A further advantage of the proposed procedure is its extensibility. The dimensionality reduction achieved through POD is linear and, in cases where the microstructure exhibits strong nonlinearities, may not yield a sufficiently small-dimensional ISV vector at the macroscale (irreducibility). To address this, nonlinear dimensionality reduction methods can be employed to further encode the POD-reduced set. For instance, POD can be used in conjunction with nonlinear autoencoders and SINDy, as demonstrated in \cite{conti2024veni} and \cite{conti2022reduced}. Additionally, multiple local reduced-order bases (ROBs) can be utilized to manage non-reducible problems by decomposing them into multiple reducible ones, aided by physics-informed cluster analysis (see, e.g., \cite{brunton2016discovering} and \cite{daniel2022physics}).

Classical homogenization techniques such as Asymptotic Expansion Homogenization (AEH) often fail to account for bifurcation or localization phenomena. While AEH provides a rigorous framework for homogenizing periodic microstructures, its formulation does not accommodate bifurcation phenomena, limiting its applicability in scenarios where localization occurs. AEH is inherently restricted to smooth solutions within a standard continuum framework. However, the TANN formulation presented here is not bound to AEH or any specific homogenization technique, making it adaptable to alternative frameworks. Notably, the TANN methodology for homogeneous materials is not limited to Cauchy continua. A robust homogenization framework based on higher-order continuum theories could enable the POD-TANN workflow to model complex microstructures that account for localization and bifurcation, thereby expanding the applicability of the approach.

\subsection{Macroelement derivation, calibration and reuse}\label{S5.2}
The proposed application of the POD-TANN approach to macroelement derivation demonstrates the significant potential of the methodology. However, its current implementation exhibits limitations.

A primary limitation of the approach lies in its specificity. The entire procedure, encompassing the computational setup, data collection, preprocessing, and analysis, is highly tailored to the particular scenario under investigation. When the scenario changes - whether in terms of domain geometry, material properties, or boundary conditions - the entire computational process must be repeated. This also includes the task of training a new neural network, which is intrinsically linked to the specific problem it was trained on. Consequently, the approach is restricted in its ability to generalize across different scenarios of the same problems without substantial reconfiguration and retraining.

Traditionally, the calibration of macroelements in geomechanics has been rooted in the parametric formulation of these elements, aimed at reducing the need for extensive numerical computations by expressing macroelement behavior in terms of adjustable parameters functions of scenario-specific geometric data and material properties (see, e.g., \cite{gorini2023multiaxial}). 

In this context, Transfer Learning (TL) presents a promising strategy to extend the applicability of the POD-TANN approach across various scenarios.

Specifically, TL could enhance the POD-TANN methodology in two significant ways:
\begin{itemize}
    \item Fine-tuning pre-trained networks: TL allows for the adaptation of a pre-trained neural network to new scenarios, even when these scenarios differ in domain geometry or material parameters. The pre-trained network, which encapsulates the learned behavior from the original scenario, can be fine-tuned using a smaller, scenario-specific dataset. This approach significantly reduces the computational effort associated with training a new network from scratch for each new problem. This concept has been successfully demonstrated in recent studies, such as \cite{qu2023data}, where TL has been used to adapt models to different geotechnical conditions efficiently.
    \item Development of parametric macroelement networks: TL algorithms can be designed to link specific model input parameters, such as geometry and material properties, to each new scenario. By incorporating these parameters into the force vector and initial conditions, it is possible to develop a parametric macroelement network. Such a network would generalize across a range of scenarios, providing a computationally efficient tool for geomechanical modeling. In this perspective, \cite{zhang2023reliability} demonstrated the potential of TL in parametric modeling by applying it to the stability analysis of slopes with varying spatial distributions, highlighting the achievable efficiency gain.
\end{itemize}

An important consideration in this context is the number of POD modes used in the model, as this directly influences the dimensionality of the internal state variable vector. The selection of POD modes is crucial, as it determines how well the reduced-order model can capture the essential features of the system. For TL to be effective across different scenarios, the selected number of POD modes should be representative across a range of macro-model parameters, including variations in geometry, material properties, and boundary conditions, as stated before. This requirement could ensure the number of POD modes (i.e. of ISV vector components) to remain unchanged and sufficient for new scenarios, thereby enabling the successful application of TL, without TANN's hyperparameters changes (e.g., the number of inputs, and consequently the number and density of hidden layers).

In this view, the need to maintain a consistent number of POD modes across different scenarios could serve as an additional criterion for determining the appropriate number of POD modes. Rather than solely focusing on ensuring that the learning targets are met, this approach would also consider the need to encompass a class of possible macro-model parameters, facilitating more effective TL. 

This dual consideration could help balance the trade-off between model accuracy and computational efficiency while broadening the applicability of the POD-TANN approach.

\section{Conclusions}\label{S6}
The new POD-TANN methodology integrates Proper Orthogonal Decomposition (POD) into the training of Thermodynamics-based Artificial Neural Networks (TANN, \cite{Masi_2021_a}) to model the macroscopic behavior of inelastic systems with microstructure. 

The POD is applied to datasets of internal coordinates (IC) that describe microscale processes in heterogeneous, inelastic materials, to obtain the interal state variables (ISVs) at the macroscale. This method leverages the hierarchical structure of POD modes, ensuring that the first N modes capture the most significant information. The ISV vector's components are hierarchically sorted, allowing dimensionality reduction to be decoupled from TANN training, offering flexibility in the number of ISV components used based on the desired accuracy.

One major advantage is the ability to incorporate microscopic information into the macroscale model using a simple and robust linear dimensionality reduction technique. The invertibility of the linear mapping enables both macroscopic homogenization and microscopic field reconstruction. Depending on the objective - whether macroscopic behavior or detailed microscopic fields description - the number of POD modes can be adjusted to balance training efficiency and reconstruction accuracy.

The methodology is demonstrated through three increasingly complex applications: homogenization of inelastic RUCs with continuous microstructure and a 3D geotechnical monopile model subjected to horizontal loading. Additionally, a RUC with spatially correlated random fields of constitutive parameters is investigated in the Appendix. In each case, the trained network accurately reproduced stress-strain or force-displacement paths, and the reconstructed microscopic fields' errors were analyzed based on the number of POD modes.

The method excludes energy terms in the loss function during training, making it applicable to diverse data sources. These could include experimental data without energy measurements. 

Overall, the POD-TANN approach offers a versatile and powerful tool for modeling heterogeneous inelastic microstructured systems, ensuring thermodynamic consistency and efficient model reduction. The results are promising and suggest the potential for even more complex applications.

\section{Acknowledgment}\label{Acknow}
The authors gratefully thank the funding of the European Research Council (ERC) under the Horizon 2020 research and innovation program of the European Union (Grant agreement ID 757848 CoQuake). In addition, the authors would like to thank Dr. Filippo Masi and the whole CoQuake group for our fruitful discussions.

\newpage
\appendix

\section*{Appendix A: Dimensionality reduction of ICs via POD}\label{Appendix_A}
Mathematically, the POD allows to write a generic field $\boldsymbol{a}(\boldsymbol{x}, t)$ by means of a modal expansion (see, e.g., Lumley \cite{lumley_1967}):
\begin{equation}\label{EQ_A1}
    \boldsymbol{a}(\boldsymbol{x}, t)=\sum_{k=1}^{n}c_k(t)\boldsymbol{\Phi}_{k}(\boldsymbol{x})=\boldsymbol{\Phi}(\boldsymbol{x})\boldsymbol{c}(t),
\end{equation}
in which the column of $\boldsymbol{\Phi}$, $\boldsymbol{\Phi}_{k}\left(\boldsymbol{x}\right)$, are deterministic time-independent spatial functions (POD modes), modulated by the time-dependent coefficients $c_k(t)$, collected in the column vector $\boldsymbol{c}(t)$. The data-driven problem associated to the use of POD, is to find the optimal basis of modes,  $\boldsymbol{\Phi}$, from data of $\boldsymbol{a}$. In practice, this is achieved exploiting the Singular Value Decomposition (SVD), \cite{brunton2022data}. 

In the following, $\boldsymbol{A}$ is used to denote a generic snapshot matrix, i.e., a collection of column vectors of snapshots of the fields $\boldsymbol{a}(\boldsymbol{x}, t)$, at subsequent time instants. The notation in equation \ref{EQ_A1} is thus transformed into $\boldsymbol{A}=\boldsymbol{\Phi}\boldsymbol{C}$, where $\boldsymbol{C}$ collects the time varying coefficients of each mode, each column associated to a different time instant.

The SVD is a unique matrix decomposition that exists for every complex valued matrix $\boldsymbol{A}\in \mathbb{C}^{n\times m}$, \cite{brunton2022data}:
\begin{equation}\label{EQ_A2}
    \boldsymbol{A} = \boldsymbol{U}\boldsymbol{S}\boldsymbol{V}^{*},
\end{equation}
where $\boldsymbol{U}\in\mathbb{C}^{n\times n}$ and $\boldsymbol{V}\in\mathbb{C}^{m\times m}$ are unitary matrices with orthonormal columns, and $\boldsymbol{S}\in\mathbb{R}^{n\times m}$ is a matrix with real non-negative entries in the diagonal and zeros elsewhere, with $(*)$ the complex conjugate transpose. 

When $n>m$, $\boldsymbol{S}$ has at most $m$ non-zero elements. The following exact economy representation of $\boldsymbol{A}$ is therefore possible:
\begin{equation}\label{EQ_A3}
    \boldsymbol{A}=\left[ \boldsymbol{\hat{U}}\quad \boldsymbol{\hat{U}}^{\perp}\right] \left[ \boldsymbol{\hat{S}} \quad \boldsymbol{0} \right]^{T} \boldsymbol{V}^{*}  = \boldsymbol{\hat{U}} \boldsymbol{\hat{S}} \boldsymbol{V}^{*}.
\end{equation}
The columns of $\boldsymbol{\hat{U}}^{\perp}$ span a vector space that is complementary and orthogonal to that spanned by $\boldsymbol{\hat{U}}$. The diagonal elements of $\boldsymbol{\hat{S}}$ are the singular values.

The optimal rank-$r$ approximation to $\boldsymbol{A}$, in a least squares sense, is given by the rank-$r$ SVD truncation $\boldsymbol{\tilde{A}}$, \cite{eckart1936approximation}:
\begin{equation}\label{EQ_A4}
    \tilde{\boldsymbol{A}}=\tilde{\boldsymbol{U}}\tilde{\boldsymbol{S}}\tilde{\boldsymbol{V}}^{*}=\inf_{\hat{\boldsymbol{A}}: rank(\tilde{\boldsymbol{A}})=r} \lVert \boldsymbol{A} - \hat{\boldsymbol{A}}\lVert_{F}.
\end{equation}
    
In the above equation, $\boldsymbol{\tilde{U}}$ and $\boldsymbol{\tilde{V}}$ denote the first $r$ leading columns of $\boldsymbol{U}$ and $\boldsymbol{V}$, and $\boldsymbol{\tilde{S}}$ contains the leading $r \times r$ sub-block of $\boldsymbol{S}$, $\lVert \bullet \lVert_F$ is the Frobenius norm. For a given rank $r$, there is no better approximation for $\boldsymbol{A}$, in the $\ell_2$ sense, than the truncated SVD approximation, $\tilde{\boldsymbol{A}}$. 

In the POD-TANN approach, the generic field $\boldsymbol{a}$ is specialized to the field of Internal Coordinates, $\boldsymbol{\xi}$, and the snapshot matrix $\boldsymbol{A}$ into $\boldsymbol{\Xi}$. Since the POD modes are time-invariant, the time-varying POD coefficients of the low-rank approximation of $\boldsymbol{\xi}$ may be used as a surrogate variable for tracking the irreversibility in the system. Thus, the low-rank SVD decomposition of $\boldsymbol{\Xi}$, $\tilde{\boldsymbol{\Xi}}=\tilde{\boldsymbol{U}}\tilde{\boldsymbol{S}}\tilde{\boldsymbol{V}}^{*}$, is exploited to identify the time-varying reduced order state variables of the macro-system, $\boldsymbol{C}\equiv\boldsymbol{Z}$:
\begin{equation}\label{EQ_A5}
    \boldsymbol{\Xi}\approx\tilde{\boldsymbol{\Xi}} = \tilde{\boldsymbol{U}}\tilde{\boldsymbol{S}}\tilde{\boldsymbol{V}}^{*} = \boldsymbol{\Phi} \boldsymbol{Z}\to \begin{cases}
      \boldsymbol{\Phi} = \tilde{\boldsymbol{U}}, \\
      \boldsymbol{Z} = \tilde{\boldsymbol{S}}\tilde{\boldsymbol{V}}^{*} = \tilde{\boldsymbol{U}}^{*} \tilde{\boldsymbol{\Xi}}.
    \end{cases} 
\end{equation}
The matrix $\boldsymbol{Z}$ collects the time-varying POD coefficients describing the influence of each POD mode at the considered instant of time. The columns of $\boldsymbol{Z}$, $\boldsymbol{\mathcal{Z}}$ can thus be interpreted as the ISV vectors at the subsequent time instants.

\newpage
\renewcommand{\thefigure}{B.\arabic{figure}}
\setcounter{figure}{0}

\section*{Appendix B: Performance on a Randomly Heterogeneous RUC with Spatially Correlated Constitutive Parameters}\label{Appendix_B}
Examples presented in sections \ref{S4.1} and \ref{S4.2} feature single inclusions of increasing topological complexity. A RUC with a more complex, spatially correlated random material properties is presented in this appendix. 

\subsection*{Correlated random field simulation algorithm}
The generation of a correlated random field and its subsequent scaling to specified statistical properties involves a series of steps combining spatial domain considerations with transformations in the Fourier domain.

Initially, the domain is established as a three-dimensional grid characterized by dimensions $L_x$, $L_y$, and $L_z$ and corresponding resolutions $N_x$, $N_y$, and $N_z$. The spatial resolutions, given by $dx = \frac{L_x}{N_x}$, $dy = \frac{L_y}{N_y}$, and $dz = \frac{L_z}{N_z}$, determine the discrete spacing between points in the spatial field.

Wave numbers for each dimension are calculated using the formula $k_x = 2\pi \cdot \text{FFT}(N_x, d=dx)$, and similarly for $k_y$ and $k_z$, where $\text{FFT}$ is the Fast Fourier Transform. These wave numbers are utilized to create a three-dimensional meshgrid, which serves as the foundation for the Fourier domain representation of the field. The radial wave number $k$ is computed as $k = \sqrt{\mathbf{k}_x^2 + \mathbf{k}_y^2 + \mathbf{k}_z^2}$, encapsulating the magnitude of the wave vector at each point in the Fourier space.

A Power Spectral Density (PSD) function $S(k)$ is defined to model the energy distribution across frequencies. In this application, a Gaussian decay model $S(k) = e^{-\frac{k^2}{\kappa^2}}$ is employed, where $\kappa$ modulates the spectral decay, influencing the smoothness and correlation properties of the spatial field. Figure B.15-a depicts the Gaussian form assumed in this study, with $\kappa=5$.

The procedure continues with the generation of complex Gaussian white noise $\eta(\mathbf{k})$ in the Fourier domain. The latter is then modulated by the square root of the PSD to form the Fourier transform of the random field, expressed as $\hat{f}(\mathbf{k}) = \sqrt{S(\mathbf{k})} \cdot \eta(\mathbf{k})$. The modulated noise is transformed back to the spatial domain using the inverse Fast Fourier Transform (IFFT), resulting in the correlated random field $f(\mathbf{x})$. This field exhibits spatial correlations as prescribed by the PSD. To align the statistical properties of the field with specific requirements, the mean and standard deviation of $f(\mathbf{x})$ are first calculated. The field is then scaled from its original mean $\mu$ and standard deviation $\sigma$ to the desired mean $\mu'$ and standard deviation $\sigma'$ using the formula $f'(\mathbf{x}) = \frac{f(\mathbf{x}) - \mu}{\sigma} \times \sigma' + \mu'$. This scaling ensures that the new field $f'(\mathbf{x})$ adheres to the specified mean and variance, maintaining the correlation structure imparted by the PSD. This generation strategy has been exploited to ensure periodicity of the generated field, inherited by the Fourier operator, differently from other techniques, e.g., the ones based on Choleski decomposition, see \cite{myers1989vector}.

\subsection*{Data generation}
The RUC voxels have been modeled using a Drucker-Prager constitutive model, with an additional volumetric cap, as detailed in \cite{abaqus_manual}. Spatially correlated random constitutive parameters have been assigned to the geometric field, which is derived from the coordinates of the Gauss points in the elements of the computational model.

Table 3 summarizes the statistical properties of the constitutive parameters used: $\beta$ and $d$ represent the friction angle and cohesion of the material, respectively. The parameter $K$ controls the dependence of the yield surface on the intermediate principal stress, while $E$ and $\nu$ are the elastic constants. The preconsolidation pressure is denoted by $p_0$, and the parameters $R$ and $\alpha$ further define the shape and hardening behavior of the cap. Figure B.1-b illustrates the spatial distribution of Young’s modulus assigned to the RUC realization. The dataset for training and validation was constructed by applying 20,000 random strain increments to the RUC, initially brought to a volumetric contraction of $0.1\%$. Further data augmentation was performed through dataset rotation, as described in Section \ref{S4.1.1}.

\begin{table}[]
    \label{A2.Tab1}
    \centering
    \begin{tabular}{|c|c|c|c|c|c|c|c|c|}
        \hline
        \textbf{} & $\beta$ (°) & $d$ (kPa) & $E$ (kPa) & $\nu$ (-) & $p_0$ (kPa) & $\alpha$ (-) & $R$ (-) & $K$ (-) \\
        \hline
        Mean & 40 & 15 & 18000 & 0.3 & 100 & 0.05 & 1.2 & 0.8\\
        \hline
        Std. & 0.1 & 0.5 & 5000 & 0.01 & 0.5 & - & - & - \\
        \hline
    \end{tabular}
    \caption{ Mean and standard deviation of the used constitutive parameters for the elas
tic-strain hardening plastic Drucker-Prager constitutive model with cap. Parameters without standard deviation are considered homogeneous.}
\end{table}

\begin{figure}\label{FIG_B1}
    \centering
        \includegraphics[width=1.0
    \textwidth]{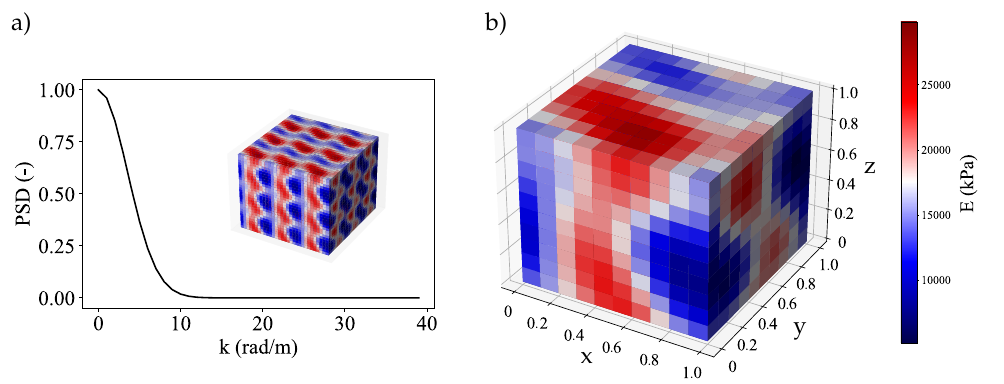}
    \caption{a) Gaussian decay used to spatially correlate the generated random field. b) Spatially correlated distribution of Young's moduli in the RUC realization.}
\end{figure}
    
\begin{figure}\label{FIG_B2}
    \centering
        \includegraphics[width=1.0
    \textwidth]{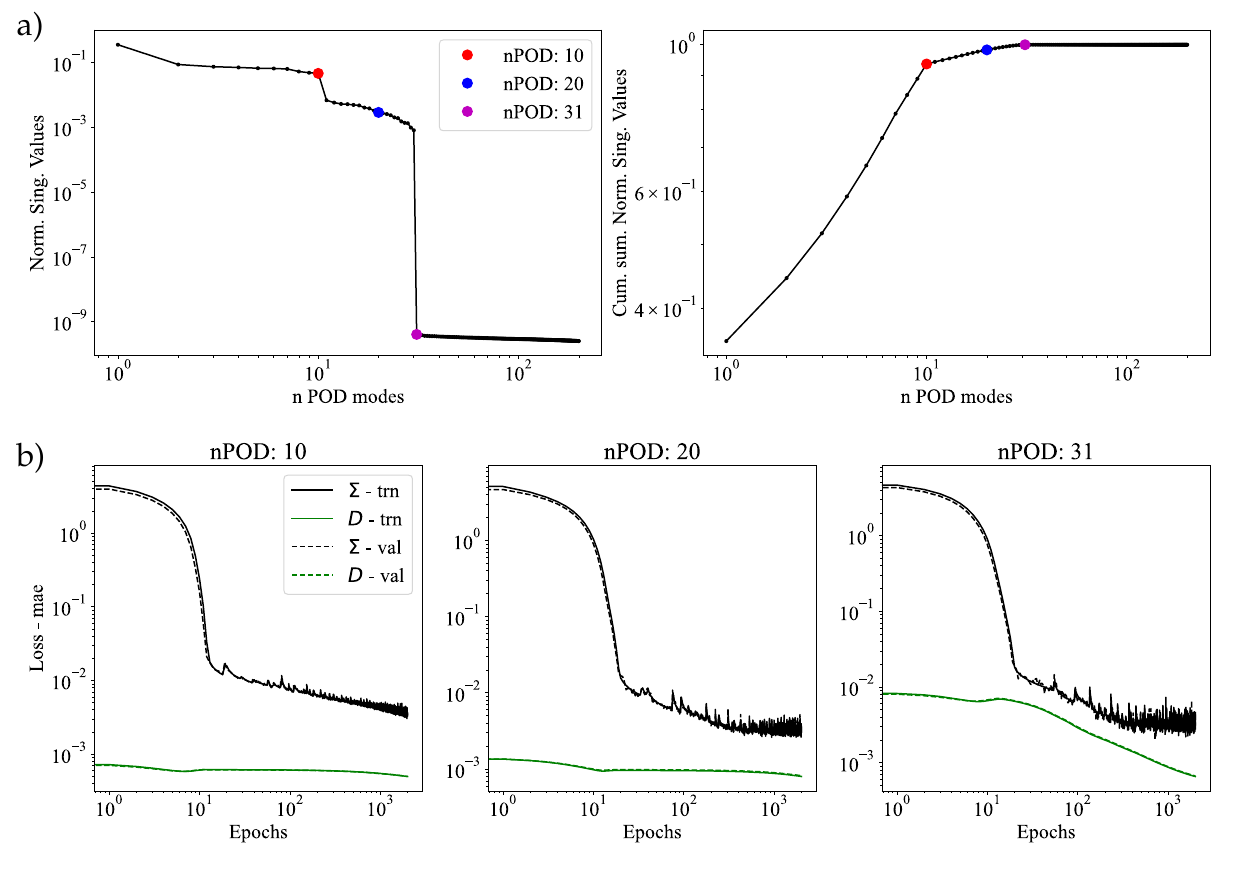}
    \caption{a) SVD decomposition of the IC set. b) Training losses considering 10, 20 and 31 POD modes.}
\end{figure}

\subsection*{Training of the TANN}
The training of the TANN energy network was guided by the analysis of the SVD of the internal coordinate dataset. Figure B.2-a presents both the normalized singular values and their cumulative sum. Three cases were considered for truncating the POD modes: 10 modes ($CR=99.9\%$), 20 modes ($CR=99.8\%$), and 31 modes ($CR=99.7\%$). The latter case corresponds to a significant drop in the relative magnitude of the singular values, marking the threshold for an almost complete representation of the original field. Figure B.2-b shows the learning curves for each case, reflecting the use of different numbers of POD modes. The hyperparameters for the energy network were inherited from those detailed in Section \ref{S4.1.2}. Training was stopped after 2000 epochs, and in all cases, the loss decreased smoothly. The final MAE reached approximately 3e-3 for stresses and 5e-4 for the rate of dissipation, consistent with the loss definition provided in Section \ref{S3.3}.  

\subsection*{Inference of the TANN}

\begin{figure}\label{FIG_B3}
    \centering
        \includegraphics[width=1.0
    \textwidth]{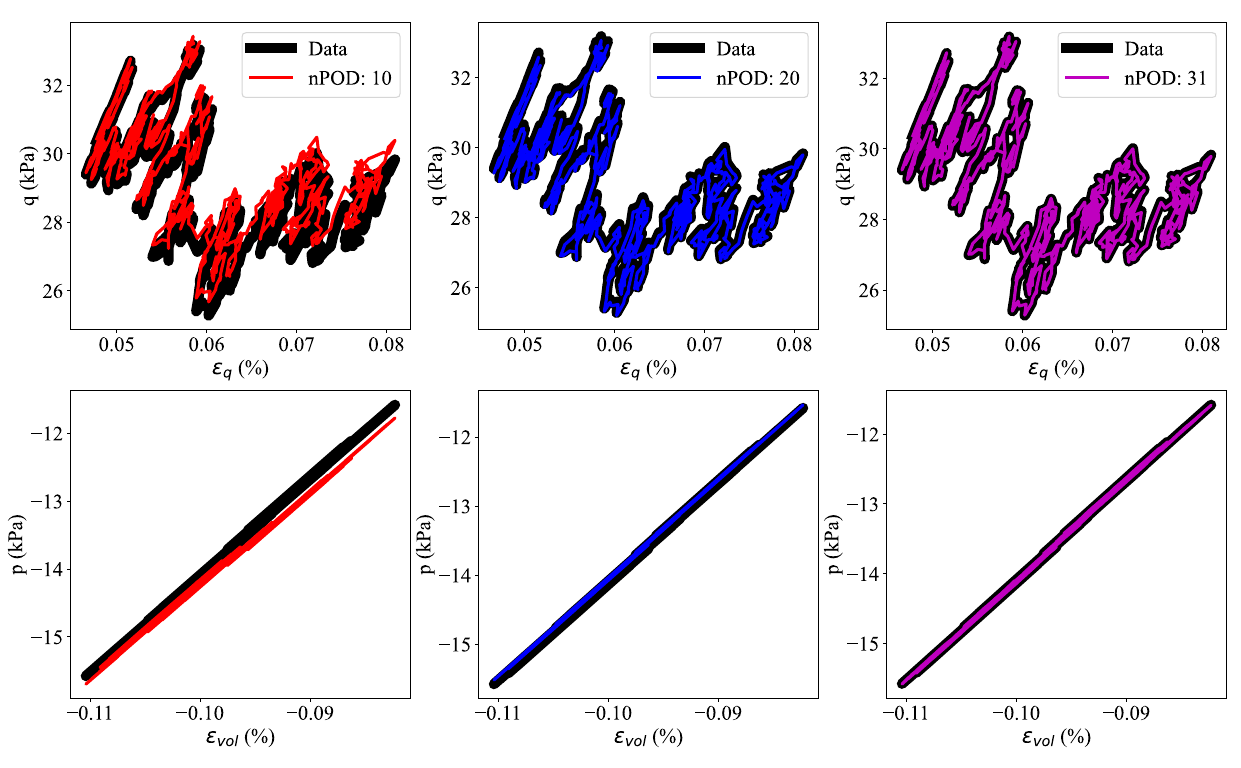}
    \caption{Comparison of the TANN inference capabilities along an unseen random path, considering 10, 20 and 31 POD modes retained.}
\end{figure}

The inference capabilities of the TANN are demonstrated on an unseen random strain path using the three trained models. Figure B.3 compares the simulated data with the TANN predictions. Although all three models reached a similar loss by the end of the training process, the accuracy in inference improves as the number of retained POD modes increases. Specifically, for the random strain path considered, the MAE is 3.21e-01 with 10 modes, 8.79e-02 with 20 modes, and 1.28e-02 with 31 modes.

\newpage
 \bibliographystyle{elsarticle-num} 
 \bibliography{cas-refs}

\begin{thebibliography}{10}
\expandafter\ifx\csname url\endcsname\relax
  \def\url#1{\texttt{#1}}\fi
\expandafter\ifx\csname urlprefix\endcsname\relax\def\urlprefix{URL }\fi
\expandafter\ifx\csname href\endcsname\relax
  \def\href#1#2{#2} \def\path#1{#1}\fi

\bibitem{raissi2019physics}
M.~Raissi, P.~Perdikaris, G.~E. Karniadakis, Physics-informed neural networks: A deep learning framework for solving forward and inverse problems involving nonlinear partial differential equations, Journal of Computational physics 378 (2019) 686--707.

\bibitem{ibanez2018manifold}
R.~Ibanez, E.~Abisset-Chavanne, J.~V. Aguado, D.~Gonzalez, E.~Cueto, F.~Chinesta, A manifold learning approach to data-driven computational elasticity and inelasticity, Archives of Computational Methods in Engineering 25 (2018) 47--57.

\bibitem{cueto2022thermodynamics}
E.~Cueto, F.~Chinesta, Thermodynamics of learning physical phenomena, arXiv preprint arXiv:2207.12749 (2022).

\bibitem{flaschel2023automated}
M.~Flaschel, S.~Kumar, L.~De~Lorenzis, Automated discovery of generalized standard material models with euclid, Computer Methods in Applied Mechanics and Engineering 405 (2023) 115867.

\bibitem{rocha2023deepbnd}
F.~Rocha, S.~Deparis, P.~Antolin, A.~Buffa, Deepbnd: a machine learning approach to enhance multiscale solid mechanics, Journal of Computational Physics (2023) 111996.

\bibitem{yin2022interfacing}
M.~Yin, E.~Zhang, Y.~Yu, G.~E. Karniadakis, Interfacing finite elements with deep neural operators for fast multiscale modeling of mechanics problems, Computer Methods in Applied Mechanics and Engineering 402 (2022) 115027.

\bibitem{Masi_2021_a}
F.~Masi, I.~Stefanou, P.~Vannucci, V.~Maffi-Berthier, Thermodynamics-based artificial neural networks for constitutive modeling, Journal of the Mechanics and Physics of Solids 147 (2021) 104277.
\newblock \href {https://doi.org/https://doi.org/10.1016/j.jmps.2020.104277} {\path{doi:https://doi.org/10.1016/j.jmps.2020.104277}}.

\bibitem{Masi_2022}
F.~Masi, I.~Stefanou, Multiscale modeling of inelastic materials with thermodynamics-based artificial neural networks (tann), Computer Methods in Applied Mechanics and Engineering 398 (2022) 115190.
\newblock \href {https://doi.org/https://doi.org/10.1016/j.cma.2022.115190} {\path{doi:https://doi.org/10.1016/j.cma.2022.115190}}.

\bibitem{Masi_ETANN}
F.~Masi, I.~Stefanou, Evolution tann and the identification of internal variables and evolution equations in solid mechanics, Journal of the Mechanics and Physics of Solids 174 (2023) 105245.
\newblock \href {https://doi.org/https://doi.org/10.1016/j.jmps.2023.105245} {\path{doi:https://doi.org/10.1016/j.jmps.2023.105245}}.

\bibitem{lumley_1967}
J.~L. Lumley, The structure of inhomogeneous turbulent flows, Atmospheric turbulence and radio wave propagation (1967) 166--178.

\bibitem{holmes2012turbulence}
P.~Holmes, J.~L. Lumley, G.~Berkooz, C.~W. Rowley, Turbulence, coherent structures, dynamical systems and symmetry, Cambridge university press, 2012.

\bibitem{brunton2022data}
S.~L. Brunton, J.~N. Kutz, Data-driven science and engineering: Machine learning, dynamical systems, and control, Cambridge University Press, 2022.

\bibitem{sampaio2007remarks}
R.~Sampaio, C.~Soize, Remarks on the efficiency of pod for model reduction in non-linear dynamics of continuous elastic systems, International Journal for numerical methods in Engineering 72~(1) (2007) 22--45.

\bibitem{kerschen2002physical}
G.~Kerschen, J.-C. Golinval, Physical interpretation of the proper orthogonal modes using the singular value decomposition, Journal of Sound and vibration 249~(5) (2002) 849--865.

\bibitem{Michel2003}
J.-C. Michel, P.~Suquet, Nonuniform transformation field analysis, International journal of solids and structures 40~(25) (2003) 6937--6955.

\bibitem{Michel2004}
J.~Michel, P.~Suquet, Computational analysis of nonlinear composite structures using the nonuniform transformation field analysis, Computer Methods in Applied Mechanics and Engineering 193~(48) (2004) 5477--5502, advances in Computational Plasticity.
\newblock \href {https://doi.org/https://doi.org/10.1016/j.cma.2003.12.071} {\path{doi:https://doi.org/10.1016/j.cma.2003.12.071}}.

\bibitem{Michel2010}
J.-C. Michel, P.~Suquet, Non-uniform transformation field analysis: a reduced model for multiscale non-linear problems in solid mechanics, in: Multiscale Modeling In Solid Mechanics: Computational Approaches, World Scientific, 2010, pp. 159--206.

\bibitem{Michel2016}
J.-C. Michel, P.~Suquet, A model-reduction approach in micromechanics of materials preserving the variational structure of constitutive relations, Journal of the Mechanics and Physics of Solids 90 (2016) 254--285.
\newblock \href {https://doi.org/https://doi.org/10.1016/j.jmps.2016.02.005} {\path{doi:https://doi.org/10.1016/j.jmps.2016.02.005}}.

\bibitem{Dvorak_1992}
G.~J. Dvorak, Transformation field analysis of inelastic composite materials, Proceedings: Mathematical and Physical Sciences 437~(1900) (1992) 311--327.

\bibitem{halphen1975materiaux}
B.~Halphen, Q.~S. Nguyen, Sur les mat{\'e}riaux standard g{\'e}n{\'e}ralis{\'e}s, Journal de m{\'e}canique 14~(1) (1975) 39--63.

\bibitem{suquet1985local}
P.-M. Suquet, Local and global aspects in the mathematical theory of plasticity, Plasticity today (1985) 279--309.

\bibitem{Coleman1967ThermodynamicsWI}
B.~D. Coleman, M.~E. Gurtin, Thermodynamics with internal state variables, Journal of Chemical Physics 47 (1967) 597--613.

\bibitem{Muschik_2008}
W.~Muschik, Survey of some branches of thermodynamics, Journal of Non-equilibrium Thermodynamics - J NON-EQUIL THERMODYN 33 (2008) 165--198.
\newblock \href {https://doi.org/10.1515/JNETDY.2008.008} {\path{doi:10.1515/JNETDY.2008.008}}.

\bibitem{miehe2002strain}
C.~Miehe, Strain-driven homogenization of inelastic microstructures and composites based on an incremental variational formulation, International Journal for numerical methods in engineering 55~(11) (2002) 1285--1322.

\bibitem{Pinho_2009}
J.~{Pinho da Cruz}, J.~Oliveira, F.~{Teixeira Dias}, Asymptotic homogenisation in linear elasticity. part i: Mathematical formulation and finite element modelling, Computational Materials Science 45~(4) (2009) 1073--1080.
\newblock \href {https://doi.org/https://doi.org/10.1016/j.commatsci.2009.02.025} {\path{doi:https://doi.org/10.1016/j.commatsci.2009.02.025}}.

\bibitem{bakhvalov2012homogenisation}
N.~S. Bakhvalov, G.~Panasenko, Homogenisation: averaging processes in periodic media: mathematical problems in the mechanics of composite materials, Vol.~36, Springer Science \& Business Media, 2012.

\bibitem{galli2020macroelement}
A.~Galli, et~al., Macroelement approaches for geotechnical problems: a promising design frame-work, Rivista Italiana di Geotecnica 54~(2) (2020) 27--49.

\bibitem{baydin2018automatic}
A.~G. Baydin, B.~A. Pearlmutter, A.~A. Radul, J.~M. Siskind, Automatic differentiation in machine learning: a survey, Journal of Marchine Learning Research 18 (2018) 1--43.

\bibitem{fresca2022pod}
S.~Fresca, A.~Manzoni, Pod-dl-rom: Enhancing deep learning-based reduced order models for nonlinear parametrized pdes by proper orthogonal decomposition, Computer Methods in Applied Mechanics and Engineering 388 (2022) 114181.

\bibitem{gavish2014optimal}
M.~Gavish, D.~L. Donoho, The optimal hard threshold for singular values is $4/\sqrt{3}$, IEEE Transactions on Information Theory 60~(8) (2014) 5040--5053.

\bibitem{he2022thermodynamically}
X.~He, J.-S. Chen, Thermodynamically consistent machine-learned internal state variable approach for data-driven modeling of path-dependent materials, Computer Methods in Applied Mechanics and Engineering 402 (2022) 115348.

\bibitem{collins1997application}
I.~Collins, G.~Houlsby, Application of thermomechanical principles to the modelling of geotechnical materials, Proceedings of the Royal Society of London. Series A: Mathematical, Physical and Engineering Sciences 453~(1964) (1997) 1975--2001.

\bibitem{abaqus_manual}
M.~Smith, ABAQUS Standard User's Manual, Version 6.9, Dassault Syst{\`e}mes Simulia Corp, United States, 2009.

\bibitem{dozat2016incorporating}
D.~Timothy, Incorporating nesterov momentum into adam, Natural Hazards 3~(2) (2016) 437--453.

\bibitem{glorot2010understanding}
X.~Glorot, Y.~Bengio, Understanding the difficulty of training deep feedforward neural networks, in: Proceedings of the thirteenth international conference on artificial intelligence and statistics, JMLR Workshop and Conference Proceedings, 2010, pp. 249--256.

\bibitem{jones2022neural}
R.~E. Jones, A.~L. Frankel, K.~Johnson, A neural ordinary differential equation framework for modeling inelastic stress response via internal state variables, Journal of Machine Learning for Modeling and Computing 3~(3) (2022).

\bibitem{matlock1970correlation}
H.~Matlock, Correlation for design of laterally loaded piles in soft clay, in: Offshore technology conference, OTC, 1970, pp. OTC--1204.

\bibitem{mayoral2005determination}
J.~M. Mayoral, J.~M. Pestana, R.~B. Seed, Determination of multidirectional py curves for soft clays, Geotechnical testing journal 28~(3) (2005) 253--263.

\bibitem{yang2020thermodynamic}
G.~Yang, B.~Bai, A thermodynamic model to simulate the thermo-mechanical behavior of fine-grained gassy soil, Bulletin of Engineering Geology and the Environment 79 (2020) 2325--2339.

\bibitem{conti2024veni}
P.~Conti, J.~Kneifl, A.~Manzoni, A.~Frangi, J.~Fehr, S.~L. Brunton, J.~N. Kutz, Veni, vindy, vici: a variational reduced-order modeling framework with uncertainty quantification, arXiv preprint arXiv:2405.20905 (2024).

\bibitem{conti2022reduced}
P.~Conti, G.~Gobat, S.~Fresca, A.~Manzoni, A.~Frangi, Reduced order modeling of parametrized systems through autoencoders and sindy approach: continuation of periodic solutions, arXiv preprint arXiv:2211.06786 (2022).

\bibitem{brunton2016discovering}
S.~L. Brunton, J.~L. Proctor, J.~N. Kutz, Discovering governing equations from data by sparse identification of nonlinear dynamical systems, Proceedings of the national academy of sciences 113~(15) (2016) 3932--3937.

\bibitem{daniel2022physics}
T.~Daniel, F.~Casenave, N.~Akkari, A.~Ketata, D.~Ryckelynck, Physics-informed cluster analysis and a priori efficiency criterion for the construction of local reduced-order bases, Journal of Computational Physics 458 (2022) 111120.

\bibitem{gorini2023multiaxial}
D.~N. Gorini, L.~Callisto, A multiaxial inertial macroelement for deep foundations, Computers and Geotechnics 155 (2023) 105222.

\bibitem{qu2023data}
T.~Qu, J.~Zhao, S.~Guan, Y.~Feng, Data-driven multiscale modelling of granular materials via knowledge transfer and sharing, International Journal of Plasticity 171 (2023) 103786.

\bibitem{zhang2023reliability}
S.~Zhang, L.~Ding, M.~Xie, X.~He, R.~Yang, C.~Tong, Reliability analysis of slope stability by neural network (nn), principal component analysis (pca), and transfer learning (tl) techniques, Journal of Rock Mechanics and Geotechnical Engineering (2023).

\bibitem{eckart1936approximation}
C.~Eckart, G.~Young, The approximation of one matrix by another of lower rank, Psychometrika 1~(3) (1936) 211--218.

\bibitem{myers1989vector}
D.~E. Myers, Vector conditional simulation, in: Geostatistics: Proceedings of the Third International Geostatistics Congress September 5--9, 1988, Avignon, France, Springer, 1989, pp. 283--293.

\end{thebibliography}





\end{document}